\documentclass[lettersize,journal]{IEEEtran}
\usepackage{amsmath,amsfonts}
\usepackage{xcolor}
\usepackage{multirow}
\usepackage{algorithmic}
\usepackage{algorithm}
\usepackage{amsmath,amssymb}
\usepackage{array}
\usepackage{subcaption}
\usepackage{dirtytalk}
\usepackage{textcomp}
\usepackage{stfloats}
\usepackage{url}
\usepackage{verbatim}
\usepackage{graphicx}
\usepackage{cite}
\usepackage{makecell}
\usepackage{orcidlink}
\usepackage{adjustbox}
\usepackage{booktabs}       
\usepackage{hyperref}
\newcommand{\nicolo}[1]{\textcolor{orange}{Nicolo: #1}}
\newcommand{\minor}[1]{\textcolor{black}{#1}}

\begin{document}


\title{Federated Unlearning: A Survey on Methods, Design Guidelines, and Evaluation Metrics\\
\vspace{0.5cm}
\large \textit{This paper has been accepted on IEEE Transactions on Neural Networks and Learning Systems}\\DOI: 10.1109/TNNLS.2024.3478334}

\author{Nicolò~Romandini* \orcidlink{0000-0002-2820-5978},~\IEEEmembership{Member,~IEEE,}
        ~Alessio~Mora \orcidlink{0000-0001-8161-1070},~\IEEEmembership{Member,~IEEE,}
        ~Carlo~Mazzocca \orcidlink{0000-0001-8949-2221},
        ~Rebecca~Montanari~\orcidlink{0000-0002-3687-0361},
        ~Paolo~Bellavista \orcidlink{0000-0003-0992-7948},~\IEEEmembership{Senior Member,~IEEE\vspace{-1cm}}
\thanks{Manuscript received February 18, 2024; revised 6 July, 2024; accepted October 6, 2024. This work was partially supported by the projects SERICS (PE00000014) and RESTART (PE0000001)  under the NRRP MUR program funded by the EU-NGEU.}%
\thanks{Nicolò Romandini, Alessio Mora, Rebecca Montanari, and Paolo Bellavista are with the Department of Computer Science and Engineering (DISI), University of Bologna, Bologna, Italy (e-mail: \{name.surname\}@unibo.it).\\
Carlo Mazzocca is with the Department of Computer and Electrical Engineering and Applied Mathematics (DIEM), University of Salerno, Salerno, Italy (email: cmazzocca@unisa.it).
}%
\thanks{Asterisk
indicates the corresponding author}.}


\markboth{Journal of \LaTeX\ Class Files,~Vol.~14, No.~8, August~2021}%
{Shell \MakeLowercase{\textit{et al.}}: A Sample Article Using IEEEtran.cls for IEEE Journals}


\maketitle
\begin{abstract}
Federated Learning (FL) enables collaborative training of a Machine Learning (ML) model across multiple parties, facilitating the preservation of users' and institutions' privacy by \minor{maintaining} data stored locally. Instead of centralizing raw data, FL exchanges locally refined model parameters to build a global model incrementally. While FL is more compliant with emerging regulations such as the European General Data Protection Regulation (GDPR), ensuring the \textit{right to be forgotten} in this context — allowing FL participants to remove their data contributions from the learned model — remains unclear. In addition, it is recognized that malicious clients may inject backdoors into the global model through updates, e.g., to generate mispredictions on specially crafted data examples. 
Consequently, there is the need for mechanisms that can guarantee individuals the possibility to remove their data and erase malicious contributions even after aggregation, without compromising the already acquired \say{good} knowledge. This highlights the necessity for novel Federated Unlearning (FU) algorithms, which can efficiently remove specific clients' contributions without full model retraining.
This survey provides background concepts, empirical evidence, and practical guidelines to design/implement efficient FU schemes. Our study includes a detailed analysis of the metrics for evaluating unlearning in FL and presents an in-depth literature review categorizing state-of-the-art FU contributions under a novel taxonomy. Finally, we outline the most relevant and still open technical challenges, by identifying the most promising research directions in the field. 

\end{abstract}


\section{Introduction}
To achieve high-performance levels suitable for real-world applications, Machine Learning (ML) and, in particular, Deep Learning (DL) models need to access large amounts of data. In numerous scenarios, users willingly furnish their data to service providers for enhanced service quality and improved user experience. Practical examples include the training of natural language processing models on datasets derived from Amazon reviews \cite{10.1145/2507157.2507163} or the development of micro-video recommendation systems leveraging user data from platforms such as TikTok and Instagram \cite{10.1145/3477495.3532027}. Data analytics promise to deliver unprecedented value to companies and organizations, e.g., as pointed out by its key role in influencing election campaigns in the so-called Facebook-Cambridge Analytica scandal \cite{fbcambridge}.

Therefore, the growing outsourcing of user information has also increased interest in privacy preservation measures. In this direction, recent years have witnessed a worldwide development and establishment of privacy-protection regulations, aiming to address the growing sensibility of individuals to the treatment of their personal information by third parties. The European Union (EU) and California state have been among the pioneers in proposing such regulations, respectively with the General Data Protection Regulation (GDPR) and the California Consumer Privacy Act (CCPA). Among the enforced guidelines about the collection, storage, and processing of personal data, both GDPR and CCPA include the on-demand \textit{removal} of personal data previously voluntarily disclosed by the users. For example, the right of erasure, also known as \textit{the right to be forgotten}, is one of the cardinal articles in the EU GDPR \cite{gdpr2016} and states that \say{\textit{the data subject shall have the right to obtain from the controller the erasure of personal data concerning him or her without undue delay [...] where the data subject withdraws consent on which the processing is based [...]}.}. Similarly, CCPA \cite{californiaccpa} includes \textit{the right to delete}, which empowers business users with the right to request that businesses delete the personal information they collected from them and tell their service providers to do the same.

In the context of ML, the right to have data removed can be realized through Machine Unlearning (MU) \cite{xu2023machine}. Data owners may want to withdraw their contributions to a trained model due to privacy or security concerns \cite{7163042}. MU involves the post-processing of a trained model, by selectively eliminating the influence of specific training samples. At first glance, the intuitive approach to unlearn targeted data consists of retraining the model from scratch utilizing the remaining data. However, this seemingly straightforward method poses significant technical challenges in terms of load and performance: retraining from scratch is not only computationally expensive but also becomes impractical when faced with frequent unlearning requests. 

Furthermore, the emergence of increasingly privacy-conscious data owners and the adoption of data regulations have influenced the perception of traditional centralized ML/DL, where the training corpus frequently contains user data, e.g., habits or preferences. In this context, Federated Learning (FL) is considered a promising solution, laying the foundation for privacy-preserving ML. In FL, data remains distributed among various sources or data owners, and a global model is iteratively built by aggregating the contributions of participants (often referred to as clients), e.g.,  model updates or gradients. Consequently, clients exchange ephemeral information locally computed on their private data, which is only valid for the current state of the global ML model.   

Nevertheless, even under the FL paradigm, how to guarantee the \textit{right to be forgotten} is unclear. Indeed, the collaboratively learned model inevitably embeds the knowledge extracted from individual user data, potentially posing a risk of leaking sensitive information \cite{zhu2019deep} and vulnerability to membership attacks \cite{shokri2017membership}, i.e., the ability to infer whether a certain data sample was part of the training set. These considerations raise the following question: how should the FL system react when a client exercises \minor{their} right to be forgotten at runtime?

As for MU, the most trivial solution to erase the contribution from a specific set of data examples (e.g., the client's private data) is to exclude them from the training set in the first place. However, retraining a model from scratch with a sanitized dataset in a federated setting is often unfeasible. This could be due to the unavailability of clients that have already participated or the excessive cost in terms of required time, computation, and energy. Therefore, there is the need for effective Federated Unlearning (FU) methods, directly applicable to already trained or in-learning models, without discarding the \say{good} knowledge acquired after the unlearning point. \minor{At the beginning of 2020, only one paper \cite{liu2020learn} aimed to remove user memorization in FL. However, the interest in FU has been rapidly increasing over the years, as highlighted by the fact that more than 60\% of the papers proposing new FU methods, reviewed in this survey, were published after 2022.}


\begin{table*}[t!]
\centering
\caption{Comparison of surveys in the field.}
\label{tab:comparison}
\begin{adjustbox}{width=\textwidth,center}
\begin{tabular}{c c c c c c c c c}
\toprule
\textbf{Year} & \textbf{Reference} &  \textbf{\makecell{Objective\\Formalization}}&\textbf{\makecell{Original\\Taxonomy}} & \textbf{\makecell{Comprehensive\\ Literature Review}} & \textbf{\makecell{Experimental\\ Insights}} & \textbf{\makecell{Comprehensive\\ Metrics Analysis}} & \textbf{\makecell{In-depth \\Security \& Privacy}}& \textbf{\makecell{Challenges \& \\Future Directions}}\\
\hline
2023 & Yang and Zhao \cite{yang2023survey} & \checkmark & \checkmark & - & - & - & - & \checkmark\\
\hline
2023 & Wang et al. \cite{10148937} & - & \checkmark & - & - & - & \checkmark & - \\
\hline
2023 & Liu et al. \cite{liu2023survey} & - & \checkmark & \checkmark & - & - & - & \checkmark\\
\hline
Now & \textbf{Ours}& \checkmark & \checkmark & \checkmark & \checkmark & \checkmark & - & \checkmark\\ 
\hline
\end{tabular}
\end{adjustbox}
\end{table*}

\subsection{Related Surveys}

Most published \minor{surveys} primarily focus on MU \cite{xu2023machine, yang2023survey, zhang2023review, qu2023learn, nguyen2022survey}, while FU is only covered from a general perspective with a limited amount of references. 

By delving into distinct aspects of FU, a few other surveys have been proposed very recently. Wang et al. \cite{10148937} review some studies on FU by mainly concentrating on the security and privacy dimensions of FU and by paying special attention to membership inference attacks and corresponding countermeasures. Liu et al. \cite{liu2023survey} present a comprehensive survey that highlights challenges, methods, and promising research directions in the field. The authors classify FU algorithms based on who initiates the unlearning and what needs to be forgotten. \minor{Methods are classified as either passive or active, depending on the client's role in the forgetting process. In addition, the paper highlights some open problems specific to FU, such as its emerging threats and the need to reach fairness and explainability.} 
Similarly, Yang et al. \cite{yang2023survey} direct their attention toward challenges and future research directions, but by adopting a high-level perspective and without providing in-depth technical details. The authors categorize works according to the objectives of FU, with limited practical and implementation insights about the surveyed solutions. To provide a clear and concise comparison, Table \ref{tab:comparison} compares this paper with the other few surveys already conducted in the general field of FU.

\subsection{Contributions}

This paper offers a comprehensive overview of FU tailored to both technical and non-technical audiences. In contrast to other surveys, our work goes beyond theoretical discussions, by providing readers also with practical design/implementation insights that contribute to clarifying all facets of FU. 

Specifically, to be highly accessible, we first present the fundamental concepts and principles of FU, encompassing objectives, motivations, and challenges. Then, we describe the primary emerging guidelines for the design of efficient FU algorithms, by including a comprehensive analysis of the metrics to assess unlearning - a critical concern within federated scenarios. Notably, we complement our reported analysis with a series of experiments that offer tangible insights into the necessity of FU. 
Moreover, we conduct an extensive review of the existing FU literature, categorizing papers through a novel taxonomy based on their objectives and the metrics employed. Finally, we identify open problems and discuss potential future research directions. We believe that this survey serves as a valuable resource for researchers seeking a clear understanding of FU and the recent related research advancements and as a practical guide for efficiently designing and implementing novel FU solutions. 
The paper's contributions can be summarized as follows:
\begin{itemize}
    \item We introduce the background concepts and the motivations for FU algorithms. We formalize the setting and possible objectives of FU methods, and what differentiates them from classical MU.
    \item We perform a set of experiments to empirically demonstrate that an FL global model memorizes contributions of individual clients' local data. Our findings show that FL global models perform significantly better on seen client data and that such knowledge fades away very slowly even if the specific clients do not participate in the next rounds. 
    \item We provide comprehensive guidelines for the efficient design and implementation of FU algorithms. This includes the requirements that the FU algorithms must meet and a detailed analysis of the metrics adopted for evaluation.
    \item We conduct a thorough review of the existing FU literature, by presenting the most relevant and updated contributions to the field, with a technical focus on the rationale behind their design choices. For ease of understanding, we classify the considered work by using a novel taxonomy presented in Section V.
    \item We identify and discuss the most relevant and open technical challenges in the field. This also outlines the most promising future research directions, by providing a roadmap for ongoing and upcoming investigations.
\end{itemize}

\subsection{Organization}

\begin{figure}[!t]
\centering
\resizebox{\columnwidth}{!}{\includegraphics{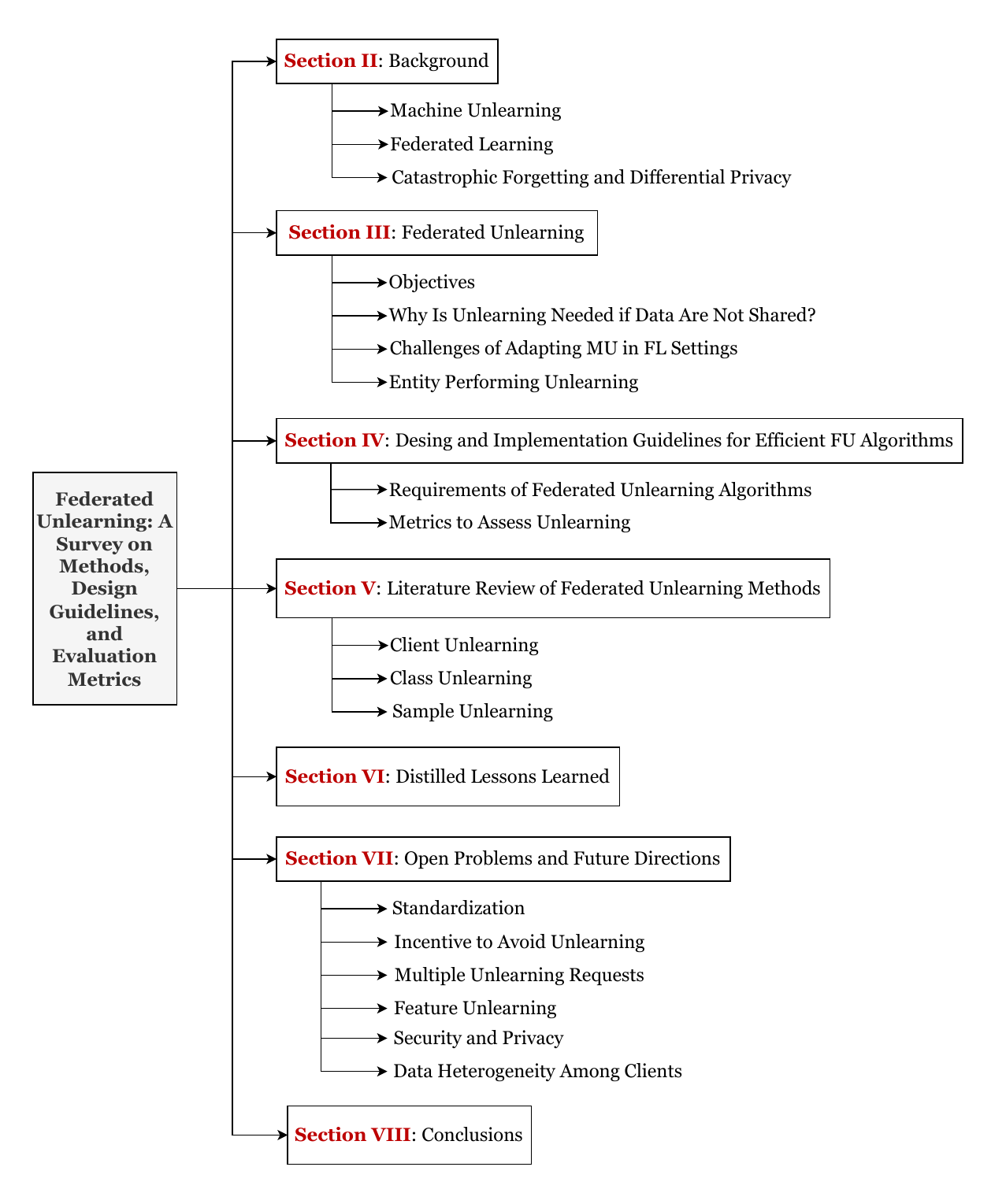}}
\caption{Illustrative organization of the paper.}
\label{fig:organization}
\end{figure}

The remainder of this paper is organized as follows. Section \ref{sec:background} introduces the needed background about MU and FL. In Section \ref{sec:fu}, we formalize the objectives of FU, we provide qualitative experimental evidence that motivates the research in the field, and we clarify the technical motivations why FU is more challenging than regular MU. In Section \ref{sec:desgud}, we offer a detailed analysis of the requirements that FU algorithms have to meet and the associated assessment metrics. Furthermore, Section \ref{sec:fu_literature} presents an in-depth exploration of all the surveyed solutions, by highlighting the rationale behind their algorithm design and implementation choices. Section \ref{sec:lesson} and \ref{sec:future_directions} respectively summarize the key lessons learned and the most promising directions for future work in FU. \minor{Finally}, Section \ref{sec:conclusion} concludes the manuscript. The structure of this survey is highlighted in Figure \ref{fig:organization}, \minor{while Table \ref{table:notation} reports the notation adopted throughout the paper.}

\begin{table}[t!]

\begin{center}
{
\caption{Notation.}
\label{table:notation}

\begin{tabular}{l c}
Symbol & Description \\
\midrule
$u$ & Target client that requests the unlearning process \\
$k$ & A client in the federation \\
$K$ & Amount of clients selected per round \\
$n$ & Total amount of clients' data examples \\
$c$ & A specific class of the classification task\\
$n_k$ & Amount of client $k$'s data samples \\
$n^c_k$ & Amount of client $k$'s data samples labeled with $c$ \\
$t$ & A round of FL training \\
$D$ & A general dataset \\
$x$ & Data sample \\
$y$ & Data sample's label \\
$L$ & Number of total classes in classification task \\
$D_u$ & Client $u$'s private dataset \\
$S_u$ & Subset of client $u$'s private dataset \\
$w$ & The weights or parameters of an ML/DL model \\
$w_t$ & Global model at a certain round $t$ \\
$w^u_t$ & Unlearned global model at round $t$ \\
$w^r_t$ & Retrained global model at round $t$ \\
$\mathcal{U}(\cdot)$ & An unlearning function or algorithm\\
$C$ & Set of training data belonging to the same class $c$\\
$g(\cdot)$ & A readout function \\
$|\cdot|$ & Cardinality of a dataset \\
$T$ & A time measurement \\
\bottomrule
\end{tabular}
}
\end{center}
\end{table}

\section{Background}
\label{sec:background}
\subsection{Machine Unlearning}
Considering a set of model weights $w$ trained on a dataset $D$, the purpose of MU is to remove the influence of a specific $D$'s subset on $w$. 
Let $D_u$ be the subset of $D$ to be forgotten, and let $D_r$ be its complement, i.e., $D_u$:$=D_{r}^\complement$ $=$ $D/D_r$. We will refer to $D_u$ as the \textit{forgetting dataset} and to $D_r$ as the \textit{retain dataset}. Let $w^r$ be a set of model weights trained only on $D_r$, usually referred to as the \textit{retrained model}. The goal of an MU algorithm is to efficiently obtain $w^{\bar{u}}$, a sanitized version of $w$, which is indistinguishable or approximately indistinguishable from $w^r$ \cite{golatkar2020eternal,xu2023machine}. Two sets of model weights are indistinguishable if an attacker with access to $w^{\bar{u}}$ cannot design a function $g(\cdot)$ to extract information specific to $D_u$. Considering, for example, a Kullback-Leibler (KL) divergence as a measure for similarity among distributions (distributions of weights in this case), the unlearned model weights $w^{\bar{u}}$ and the retrained model weights $w^r$ are formally indistinguishable if their KL divergence is zero \cite{golatkar2020eternal}:
\begin{equation}
KL(g(w^{\bar{u}}), g(w^r)) = 0    
\end{equation}
with $w^{\bar{u}} = \mathcal{U}(w)$ and $\mathcal{U}(\cdot)$ representing an unlearning algorithm. Similarly, the unlearned model weights $w^{\bar{u}}$ and the retrained model weights $w^r$ are approximately (or $\epsilon$-indistinguishable) if their KL divergence is lower than a small threshold value $\epsilon$,
\begin{equation}
KL(g(w^{\bar{u}}), g(w^r)) < \epsilon.    
\end{equation}
As anticipated, the naive solution to unlearn the contribution of $D_u$ would consist of retraining the model from scratch only on $D_r$. However, this method can be costly in terms of time, computation, and energy consumption, especially when thinking of large models. Hence, MU mechanisms should be more efficient than the retraining strategy. Furthermore, in some scenarios, the entire corpus of $D_r$ may not be directly accessible or may not be available anymore after the training of the original model, i.e., as we will see in the following sections, for FL settings. 

It is worth noting that MU mechanisms are not only needed to meet regulations about individuals' privacy, such as GDPR and CCPA, but also represent a security tool for the provider of the ML service. Training data may include natural or malicious outliers, which can harm the model performances or introduce potential security threats (e.g., model backdoor). Under this perspective, MU algorithms represent a mechanism to selectively remove the influence of such a subset of threatening data without discarding the useful knowledge already acquired. 

\subsection{Federated Learning}

Considering $K$ learners (also referred to as clients) participating in the collaborative training, with each client $k$ possessing a local dataset $D_{k}$, FL aims to minimize a global objective function, $f(w)$, that can be expressed as follows:
\begin{equation}
\label{eq:fl}
\min_{w} f(w) := \sum_{k=1}^{K}\frac{n_k}{n} F_k(w), 
\end{equation}
where $f(w)$ is a weighted average of local objective functions $F_k$, such as cross-entropy loss for a supervised classification task, across all the datasets of clients. $w$ represents the parameters of the global model, $n_k$ represents the number of training examples contained in local datasets $D_k = \{x_i, y_i\}^{n_k}_i$ with $x$ and $y$ being the data point and the related label. $n$ is the total number of data points in the federation of clients. 

Federated Averaging \cite{mcmahan2017communication}, also known as FedAvg, serves as the foundational algorithm for FL and can be described as a synchronous protocol that adopts a client-server paradigm \cite{bellavista2021decentralised}. FedAvg proceeds in rounds and heuristically tries to solve the optimization problem in Eq. \ref{eq:fl}. During each round, the server distributes the current global model to a random selection of available clients, 
which fine-tune the received model parameters on local data. Then, the \minor{selected} clients communicate a model update to the server (i.e., the difference between the locally fine-tuned and the received model parameters). Once the server has gathered such model updates, a weighted-average aggregation is performed, and the resultant averaged update is applied to the current global model. The initial formulation of FedAvg applies the resultant aggregated update by just summing it to the server model. However, aggregated updates can be seen as pseudo-gradients, and can be applied by using an optimizer of choice, i.e., different from regular Stochastic Gradient Descent (SGD) \cite{reddi2020adaptive}. 

While FedAvg is considered the baseline for FL, peer-to-peer and/or asynchronous variants exist. In peer-to-peer FL (or fully decentralized), clients directly communicate with each other without the need for a central coordinator, while in asynchronous FL the server can aggregate the contributions of clients as soon as it receives them (without waiting for the slower learners) \cite{bellavista2021decentralised}.

It is worth noting that, while sharing model updates in place of raw data makes it harder for attackers to extract information about clients' training data, FL does not ensure complete privacy protection. Indeed, it has been proved that disclosing model updates or gradients can expose clients to information leakage (e.g., \cite{zhu2019deep, melis2019exploiting, nasr2018comprehensive}).

\subsection{Catastrophic Forgetting and Differential Privacy}\label{sec:background_dp_cf}
\minor{In this subsection, we introduce catastrophic forgetting and Differential Privacy (DP), two concepts correlated with FU. In particular, we clarify why these concepts are orthogonal to unlearning.}
\\\\
\textbf{Catastrophic forgetting.} In the field of Continual Learning (CL) \cite{lesort2020continual}, a key challenge is catastrophic forgetting. This occurs when a neural network while adapting to new tasks, tends to disrupt the representations learned to solve the previous downstream tasks \cite{kemker2018measuring}. Recent studies have demonstrated that catastrophic forgetting also happens in FL training due to the round-by-round shift of data distributions. The pools of selected participants can change at each round, potentially including clients with skewed data distribution concerning the global data distribution. This can lead the global model to specialize mainly on the specific per-round data distribution of activated clients while struggling to generalize overall \cite{legate2023re, caldarola2022improving, lee2021preservation}.


Catastrophic forgetting may seem a natural method for unlearning, i.e., the global model forgets the specific contributions of a client's data as soon as th
at client is \minor{no longer} included in FL rounds. However, it has been demonstrated \cite{gao2022verifi} (see also Subsection \ref{subsec:why_unl}) that even though a participant leaves the training, the global model still retains contributions learned from its data. Consequently, relying on catastrophic forgetting cannot be considered a reliable strategy.
Moreover, when unlearning methods are applied to an ML model, there is the risk that catastrophic forgetting is induced by the unlearning, i.e., the unlearning algorithm also removes to-retain knowledge as a side effect.
\\\\
\textbf{Differential Privacy.} Within the context of ML, it is \minor{crucial} that the public release of model parameters does not leak \textit{too much} sensitive information about the original training data, even \minor{if an} adversary can have access to auxiliary information. DP  \cite{dwork2011differential, dwork2014algorithmic} \minor{addresses this challenge 
by ensuring } that the inclusion, or exclusion, of an individual's data in a private database does not \textit{essentially} affect the outcome of any analysis conducted on that database. Therefore, an individual should be \textit{almost} indifferent between disclosing her data and not\footnote{If the inclusion (or exclusion) of an individual's data did not affect the outcome at all, it would mean that the individual's data is useless.}. In a nutshell, DP is a \minor{rigorous mathematical framework to quantify privacy guarantees, formalizing the vague terms of}
 \textit{too much}, \textit{essentially}, and \textit{almost} thanks to a privacy measure \minor{called \textit{privacy budget}, denoted as} $\delta$. When applied to ML/DL models, DP injects $\epsilon$-calibrated noise, e.g., drawn from a Gaussian distribution, to perturb model updates or gradients. \minor{This noise introduces a trade-off, as it can degrade model performance} (e.g., \cite{mcmahan2017learning, geyer2017differentially, bagdasaryan2019differential}), with $\epsilon$ being the parameter that tunes the noise magnitude, the smaller the $\epsilon$ the stronger the noise. In FL processes, DP should be employed from the beginning \minor{and }it cannot be seamlessly integrated into ongoing training. \minor{DP can be considered an orthogonal tool for unlearning since it masks the participation of individuals, reducing the vulnerability to membership attacks.} \minor{On the other hand, DP} \minor{does not address the need to remove harmful contributions from malicious clients}. 




\begin{figure}[!t]
\centering
\includegraphics[width=0.49\textwidth]{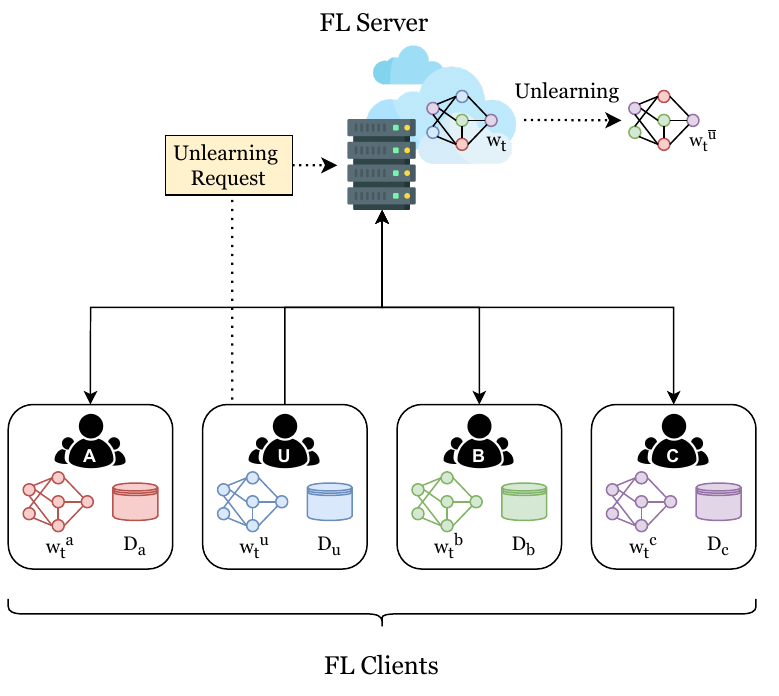}
\caption{\minor{FU} overview. When the client \minor{$u$} requests to be forgotten, the FL server erases its contributions from the global model $w_t$, generating a new version $w_t^{\bar{u}}$.  }
\label{federatedunlearning}
\end{figure}

\section{Federated Unlearning}
\label{sec:fu}
Suppose that after $t$ rounds of FL, a given client $u$ submits an unlearning request for a subset $S_u$ of its private data $D_u$ and an unlearning algorithm $\mathcal{U}$ is executed to fulfill the request. An unlearning algorithm $\mathcal{U}$ is applied to the global model $w_t$, which has been trained including client $u$ in the pool of clients, generating an unlearned global model $w^{\bar{u}}_t$, i.e., $w^{\bar{u}}_t =\mathcal{U}(w_t)$. $\mathcal{U}(\cdot)$ can be defined as a function that ensures that $w^{\bar{u}}_t$ exhibits performances indistinguishable or approximately indistinguishable from a global model $w^r_t$ trained without the contribution of $S_u$. Figure \ref{federatedunlearning} offers an illustrative overview of FU. 

FU can have different objectives (sample unlearning, class unlearning, and client unlearning), as we detail in Subsection \ref{sec:fu_objectives}. In Subsection \ref{subsec:why_unl}, we qualitatively and experimentally motivate the need for unlearning even though FL does not require direct access to raw data. In fact, the weights of the global model being trained round-by-round may leak information about clients' data. As we describe in Subsection \ref{subsec:challenges_fu}, the architectural design of FL introduces a more challenging setting than regular MU, due to the decentralization and inscrutability of data, and the iterative and stochastic nature of the FL process. In Subsection \ref{sec:requirements}, we present the requirements that FU algorithms should meet. Finally, Subsection \ref{sec:metrics} outlines and describes the metrics being used in the related literature to assess \minor{whether} an FU mechanism fulfills the previously introduced requirements. 

\subsection{Objectives}
\label{sec:fu_objectives}

\begin{figure}[!t]
\centering
\includegraphics[width=0.4\textwidth]{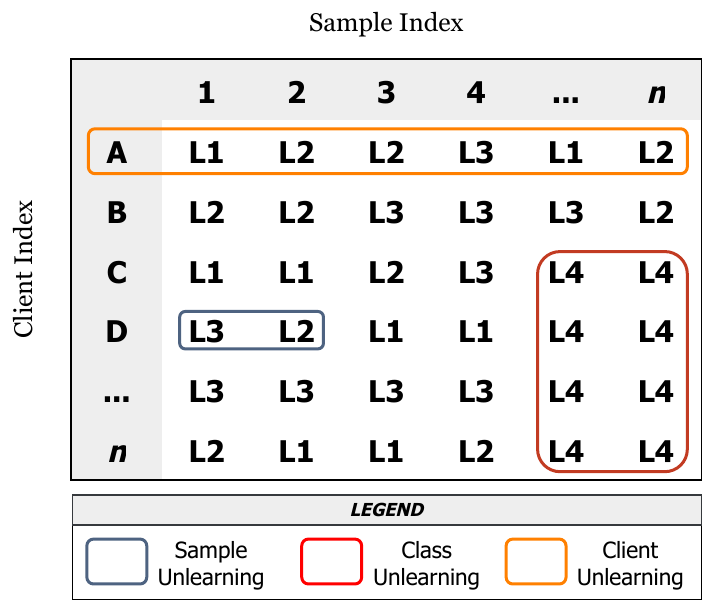}
\caption{Visualization of FU objectives. We consider four clients in the federation, i.e., clients A, B, C, and D, and four total classes for the data held by clients, i.e., L1, L2, L3, and L4. The figure reports the label that corresponds to a specific sample in the local dataset of clients (e.g., data sample 1 in client A's dataset belongs to class L1).}
\label{fig:fu_objective}
\end{figure}

As depicted in Figure \ref{fig:fu_objective}, FU aims to achieve one or more of the following unlearning objectives \cite{wu2022federated}:

\smallskip
\noindent\textbf{Sample Unlearning.} Sample unlearning aims to remove the contribution of specific data samples from the trained model. In this case, the client that requests the unlearning would like to erase the contribution of a subset of its data, 
i.e., $S_u \subset D_u$.

\smallskip
\noindent\textbf{Class Unlearning.} Class unlearning aims to remove the contribution of all the data samples that belong to a certain class $c$ across clients, i.e., to remove the contribution of $C = \bigcup_{k \in K} S_k $ with $S_k = \{x_i, c\}^{n^c_k}_i$, with $n^c_k$ the number of local samples at client $k$ labeled with class $c$. Ideally, if provided with samples belonging to a removed class, the unlearned model should produce an outcome at random between the remaining classes.

\smallskip
\noindent\textbf{Client Unlearning.} Client unlearning is peculiar to federated settings, and it relates to the \textit{right to be forgotten} of clients. In this case, the client that requests the unlearning would like to erase the contribution of its entire local dataset, i.e., $S_u = D_u$. Client Unlearning can be seen as applying Sample Unlearning on all the samples that a specific client held at the time of local training. Similarly, Client Unlearning can also be a special case of class unlearning when a single client $u$ is the only one holding all the samples for class $c$ across the federation, i.e., $C = S_u = \{x_i, c\}^{n_u}_i$ and $S_u = D_u$. 

\begin{figure*}[h!]
\centering
\begin{subfigure}[t]{.24\textwidth}
\centering
\includegraphics[width=.95\linewidth]{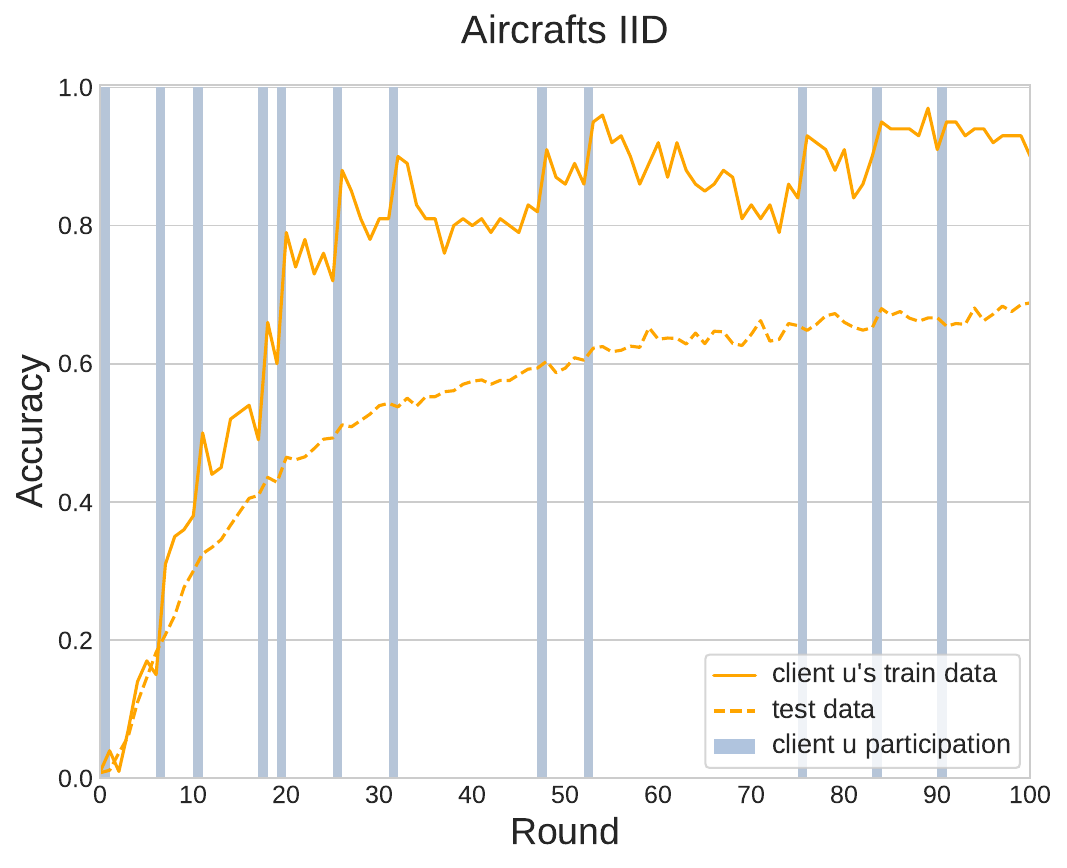}
\end{subfigure}
\hfill
 \begin{subfigure}[t]{.24\textwidth}
 \centering
  \includegraphics[width=.95\linewidth]{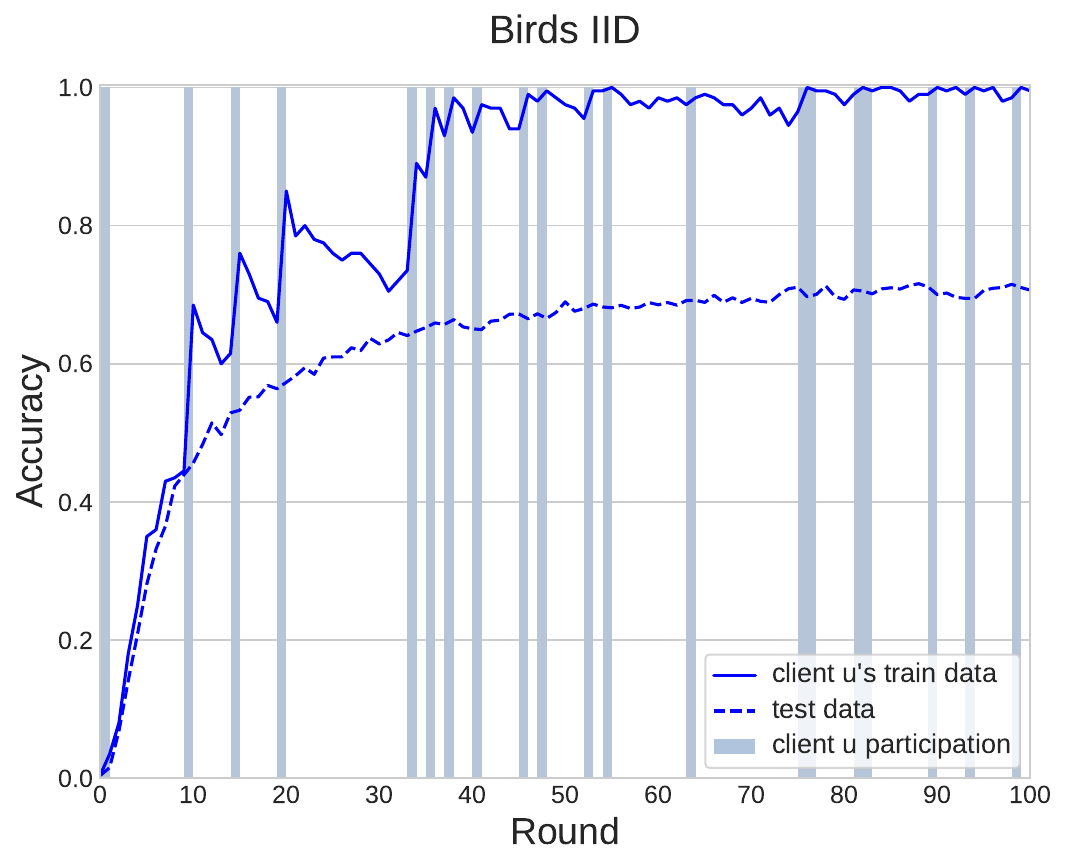}
 \end{subfigure}
 \hfill
  \begin{subfigure}[t]{.24\textwidth}
 \centering
  \includegraphics[width=.95\linewidth]{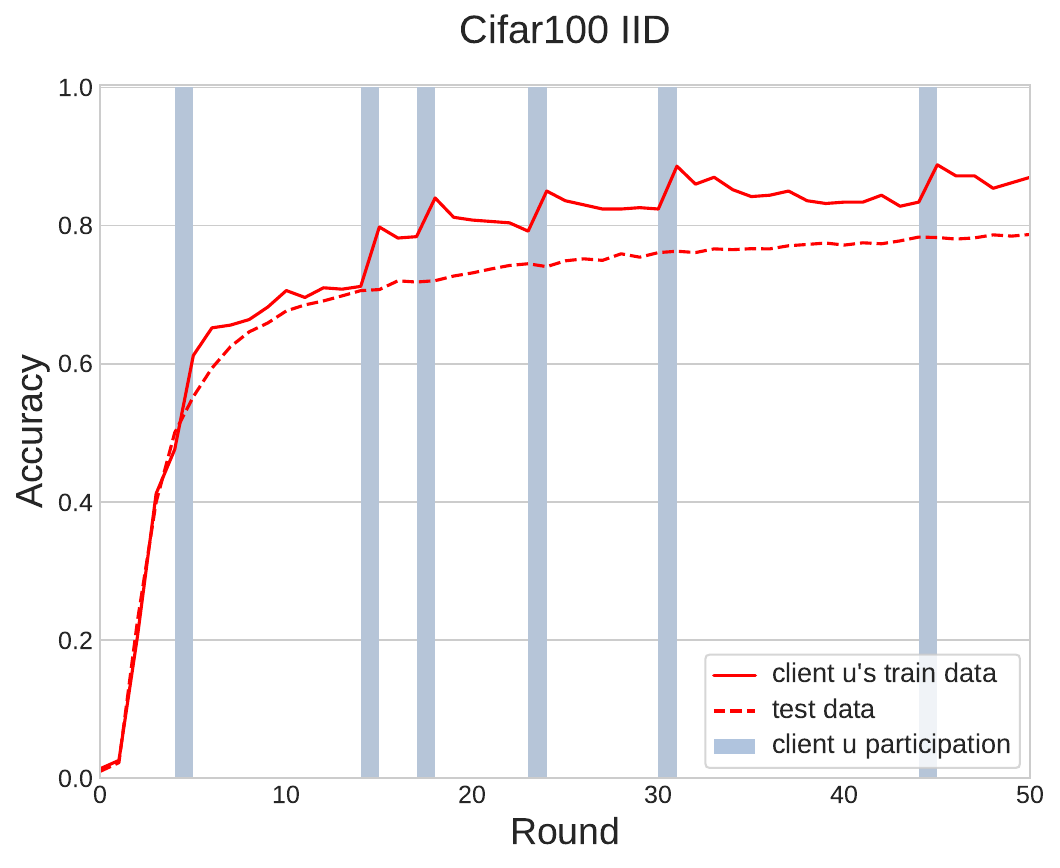}
 \end{subfigure}
  \hfill
  \begin{subfigure}[t]{.24\textwidth}
 \centering
  \includegraphics[width=.95\linewidth]{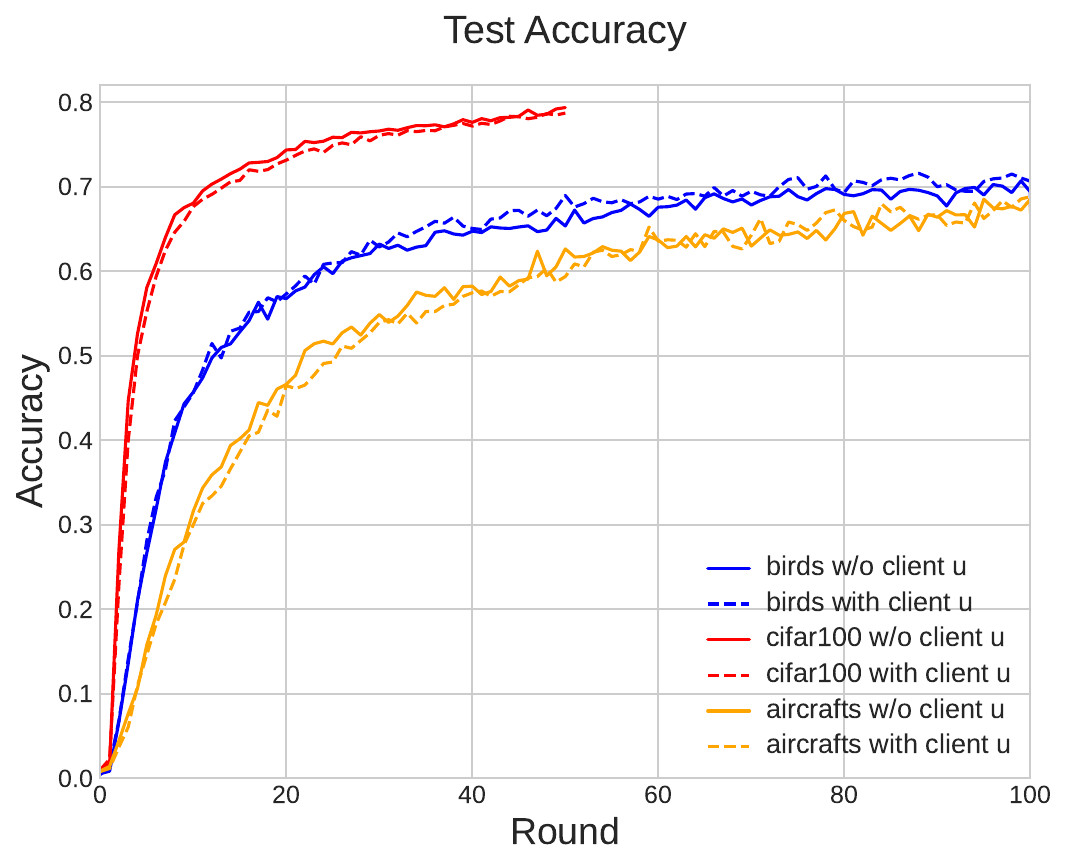}
 \end{subfigure}

\begin{subfigure}[t]{.24\textwidth}
\centering
\includegraphics[width=.95\linewidth]{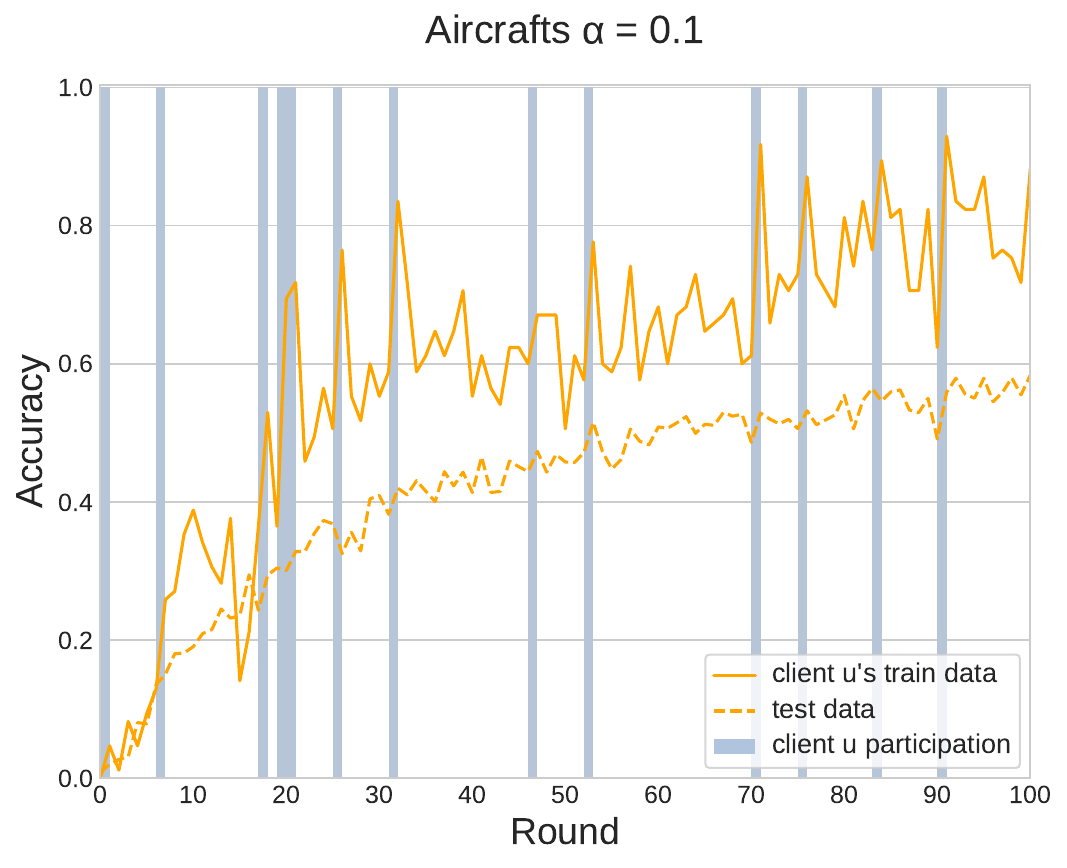}
\end{subfigure}
\hfill
 \begin{subfigure}[t]{.24\textwidth}
 \centering
  \includegraphics[width=.95\linewidth]{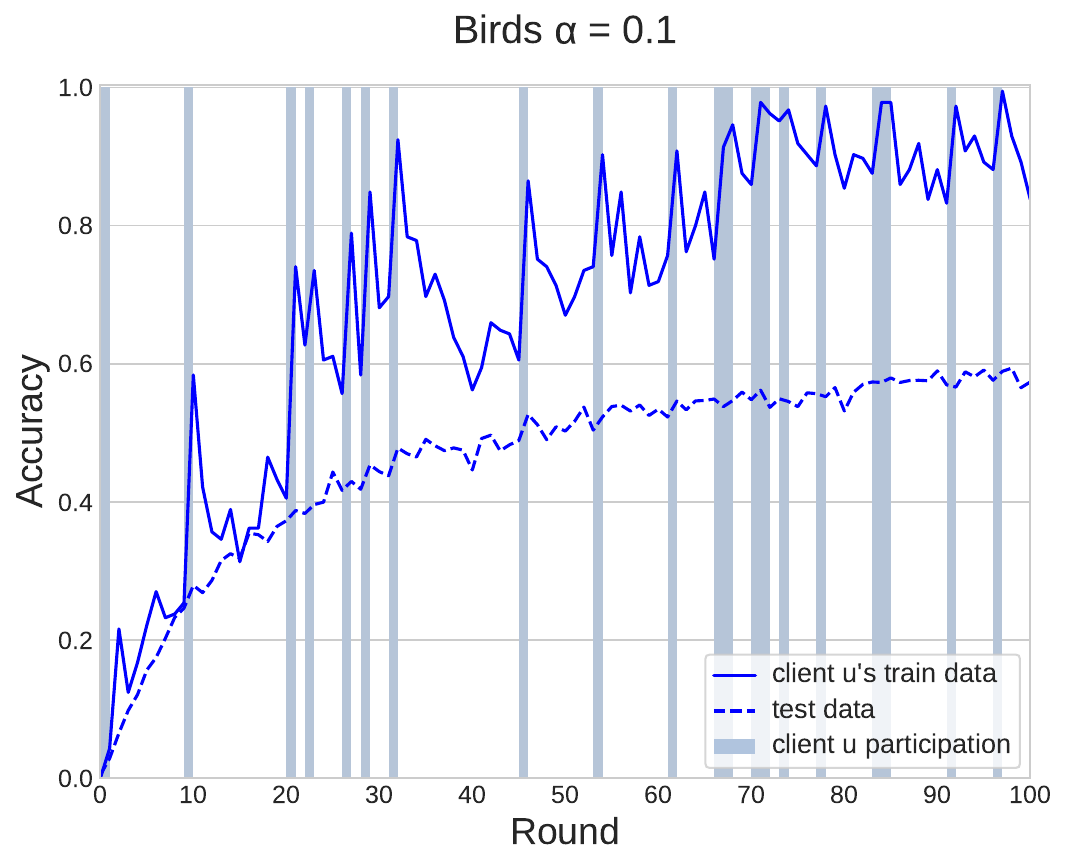}
 \end{subfigure}
 \hfill
  \begin{subfigure}[t]{.24\textwidth}
 \centering
  \includegraphics[width=.95\linewidth]{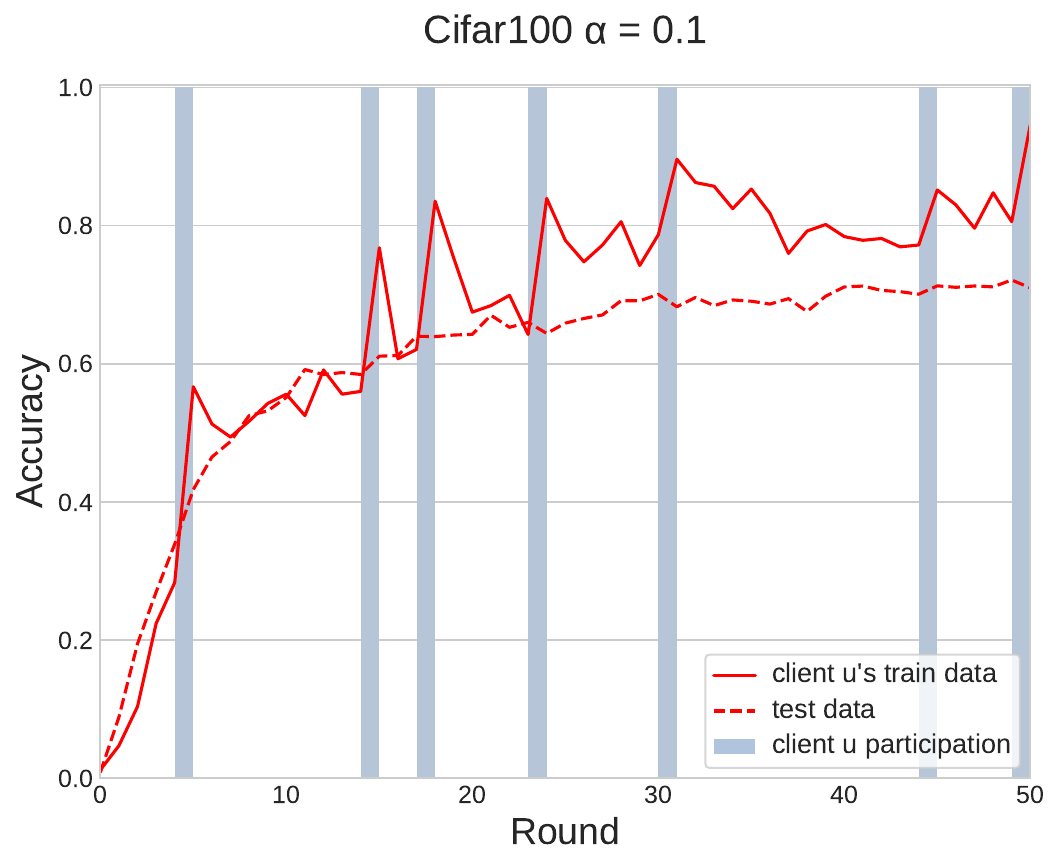}
 \end{subfigure}
  \hfill
  \begin{subfigure}[t]{.24\textwidth}
 \centering
  \includegraphics[width=.95\linewidth]{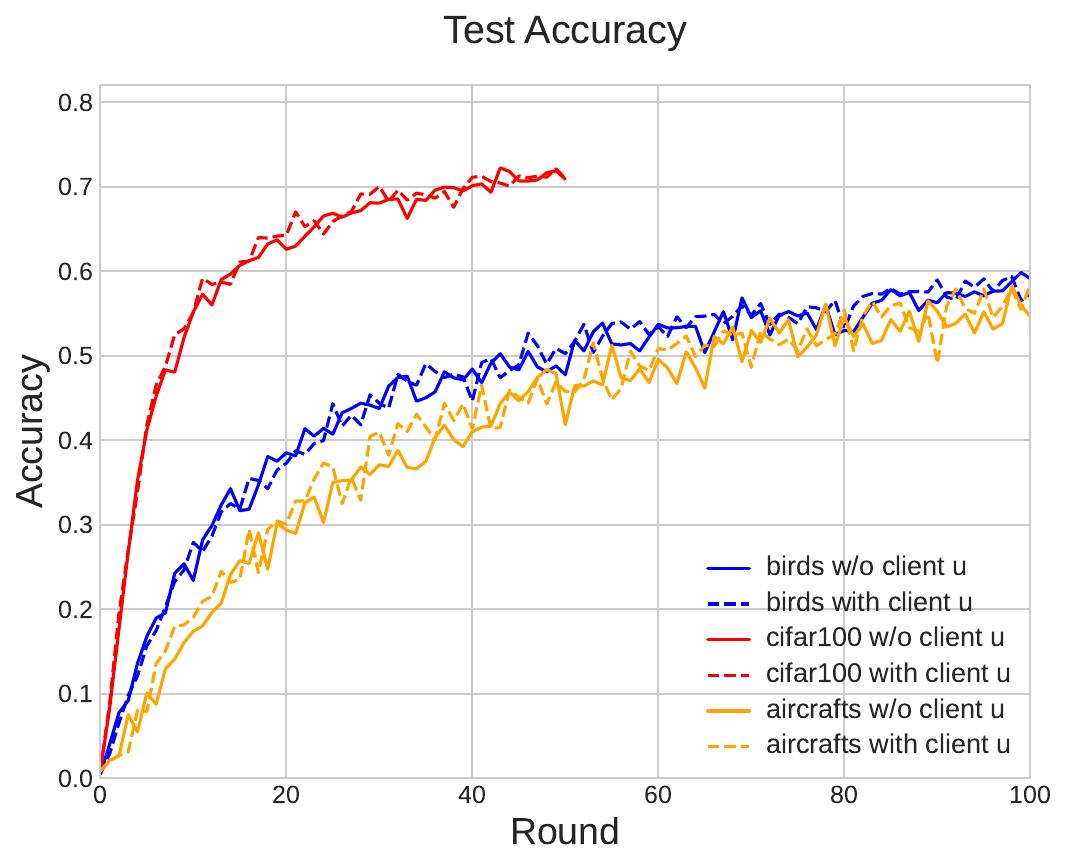}
 \end{subfigure}
\caption{The global model's accuracy on test data (solid line) versus the global model's accuracy on client $u$'s train data (dashed line) across many FL rounds. Grey bars indicate that client $u$ has participated in that round. The upper row of charts considers IID data distribution among clients. The bottom row of charts considers non-IID data distribution among clients, simulated via distribution-based labels skew (with concentration parameter $\alpha$=0.1). The last column of charts depicts the global model's test accuracy with or without client $u$ in the federation across rounds.}
\label{fig:test_vs_train_and_client_participation}
\end{figure*}

\begin{figure}[h!]
\centering
\begin{subfigure}[t]{.24\textwidth}
\centering
\includegraphics[width=.95\linewidth]{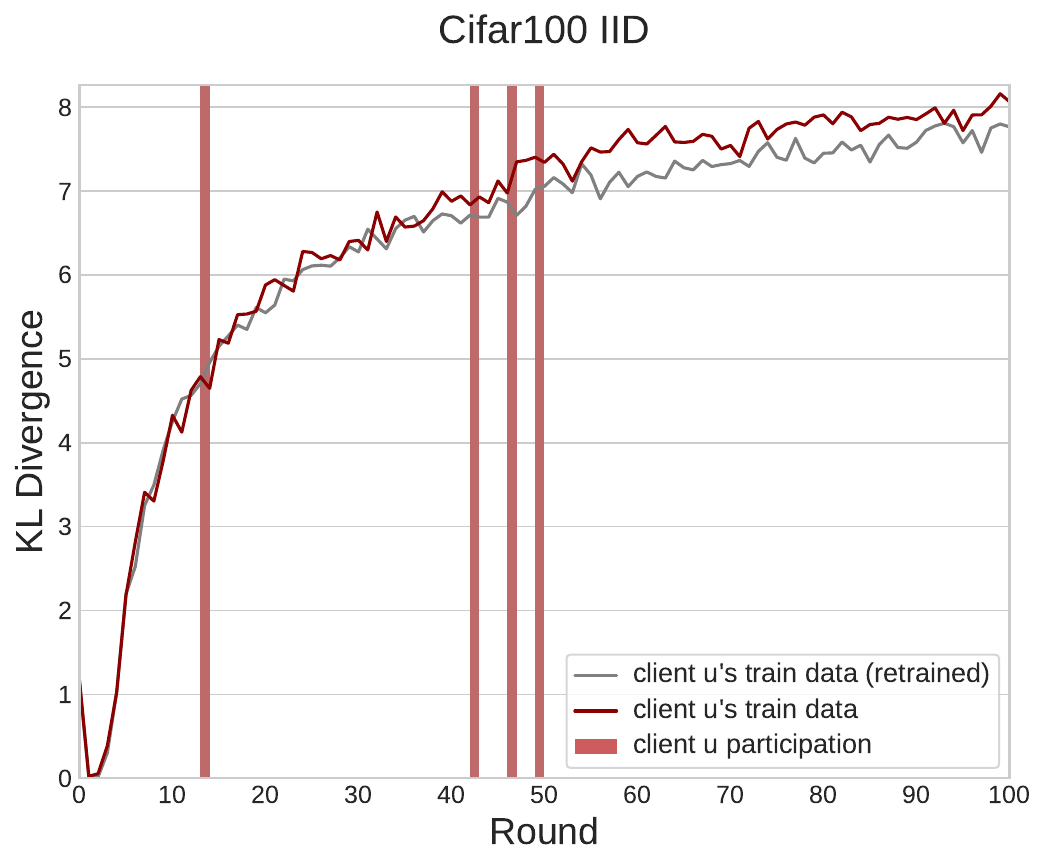}\caption{IID}
\end{subfigure}
\hfill
\begin{subfigure}[t]{.24\textwidth}
\centering
\includegraphics[width=.95\linewidth]{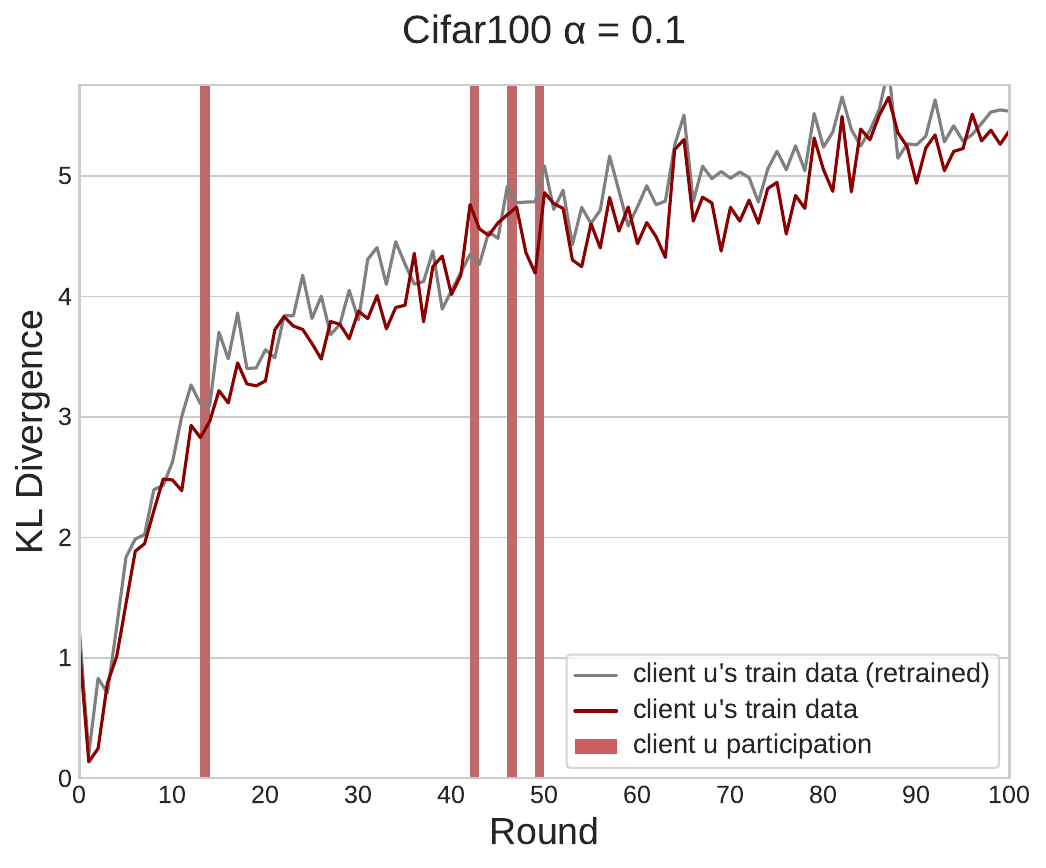}
\caption{Non-IID}
\end{subfigure}
\caption{The KL divergence between the global model's output probability on client $u$'s train data and a uniformly distributed output probability (as proposed in \cite{gao2022verifi}), across many FL rounds. The red line refers to regular training (client $u$ is included in the federation only for the first 50 rounds). Red bars indicate that client $u$ has participated in that round. The gray line refers to retraining (sanitized federation) where client $u$ never participates. }
\label{fig:natural_forgetting}
\end{figure}

\subsection{Why Is Unlearning Needed if Data Are Not Shared?}
\label{subsec:why_unl}  
To intuitively understand the impact of a client's data on the global model we performed a set of experiments, whose results are shown in Table \ref{table:results_train_test}. 

Specifically, we compared the performance in accuracy of a global model trained including or excluding a specific client $u$ in the federation. While the test accuracy is very similar among such two versions of the global model, the global model exhibits significantly higher accuracy on client $u$'s train data when the federation includes client $u$. This means that the global model trained with client $u$ among the federation may leak information about client $u$'s private data and that may be susceptible to, at least, membership attacks\footnote{As we will deepen in Sec. \ref{sec:metrics}, membership attacks tries to determine whether one or more specific samples have been used for training.}. The same trend emerges with homogeneous and heterogeneous data distributions among clients. 

Table \ref{table:results_train_test} reports the accuracy of a global model $w$ trained on a federation of participants including or excluding a specific client $u$. Three datasets for image classification were considered (CIFAR-100, Birds and Aircraft), and a visual transformer architecture was used (MiT-B0 \cite{xie2021segformer}), starting from a pretrained checkpoint and performing FL for 100 rounds. The \textit{test} column refers to the global model's accuracy on a test set that clients never see during training. The \textit{train} column refers to the global model's accuracy on the train set of client $u$. We tracked the accuracy of the global model on test and client $u$'s train data across 100 rounds when the federation includes client $u$, visually highlighting whether client $u$ participates in a round. As evident from Figure \ref{fig:test_vs_train_and_client_participation}, when client $u$ is included in the training round, the global model spikes up to a higher level of accuracy on client $u$'s train set. It is worth noting that if client $u$ does not participate for a while, the global model tends to become less accurate on that train data, but still above the test performances. This phenomenon has also been noted in \cite{gao2022verifi}. 

\begin{table}[t!]

\begin{center}
{
\caption{Accuracy (\%) of a global model trained with or without (w/o) a specific client in the federation.}
\label{table:results_train_test}

\begin{tabular}{c l | c c | c c}
\toprule
& & \multicolumn{2}{c}{$w$ trained w/o client $u$} & \multicolumn{2}{c}{$w$ trained with client $u$} \\ 
& Dataset & Test & \minor{\makecell{Client $u$\\ Train  Data}} & Test & \minor{\makecell{Client $u$\\ Train  Data}} \\
\midrule
\multirow{3}{*}{\rotatebox[origin=c]{90}{IID}} & CIFAR-100 & 79.37 & 76.80 & 78.72 & \textbf{87.00} \\
& Aircraft & 68.53 & 66.00 & 68.83 &  \textbf{90.00}\\
& Birds & 69.4 & 68.00 & 70.66 & \textbf{99.50} \\
\midrule
\multirow{3}{*}{\rotatebox[origin=c]{90}{non-IID}} & CIFAR-100 & 70.90 & 67.34 & 70.98 & \textbf{94.60} \\
& Aircraft & 54.64 & 58.82 & 58.36 & \textbf{88.24}\\
& Birds & 59.13 & 57.3 & 57.37 & \textbf{83.78} \\
\bottomrule
\end{tabular}
}
\end{center}
\end{table}

Moreover, to understand whether \textit{natural forgetting} could be a viable option for unlearning, i.e., just excluding the client that requested its removal from the federation, we performed a set of further experiments.
We want to empirically determine whether the contribution of that client naturally fades away after one or a few rounds, taking advantage of catastrophic forgetting (see also Sec. \ref{sec:background_dp_cf}). We simulated a large-scale federation with 500 clients training for 100 rounds (4\% participation rate), with client $u$ leaving at round 50. As depicted in Figure \ref{fig:natural_forgetting}a (homogeneous data among clients), after client $u$ leaves and for the rest of the FL process, the global model steadily keeps generating more meaningful predictions than the retrained baseline on client $u$'s train data for all the tracked rounds. Furthermore, the global model still outputs more accurate predictions (with lower loss as well) on client $u$'s train data compared to a sanitized federation from the beginning (see also the complete metrics in Figure \ref{fig:natural_forgetting_complete}). Contrarily, when data are heterogeneous (reported in Figure \ref{fig:natural_forgetting}b), catastrophic forgetting seems to help natural unlearning: after a few rounds, there is no sensible difference between the tracked metrics for the two versions of the global model. 

Overall, the empirical evidence shows that, even with a large-scale federation of clients holding small local datasets, the influence of a client's contribution on the global model may remain noticeable for long at least when data are homogeneously distributed.
Further details about the experimental settings, employed metrics, and reference to our code repository are provided in Appendix \ref{appendix:exp_setup}.


\subsection{\minor{Challenges of Adapting MU in FL Settings}}
\label{subsec:challenges_fu}
Concerning centralized MU, FL settings introduce even more challenging constraints. The iterative nature of collaborative learning, the stochastic selection of per-round participating clients, and the decentralization and inscrutability of local data represent additional degrees of complexity that have to be accounted for in FU algorithms.
\\\\
\textbf{Non-deterministic Training Process.} \minor{Most of the MU mechanisms suppose that retained data are available when unlearning is performed \cite{xu2023machine}. Contrarily, the data available in an FL round depends on the client selection, which is subject to various dynamics.} For example, clients eligible for a certain round may not be available in the future due to connection or battery issues, or unwillingness to participate. This peculiarity of FL makes the possibility of exactly reproducing the rounds after $t_{u}$ unrealistic. At the same time, discarding all the knowledge acquired between the aggregation at round $t_{u}$ and the aggregation at round $t_{i}$ (i.e., the most recent round), that possibly may be distant in time with $t_u<<t_i$, would waste significant learning efforts of the federation. In this sense, methods that can embed that knowledge are most appealing. 
\\\\
\textbf{Inscrutable Model Updates and Finite Memory.}
In principle, the FL aggregator or server should not or could not be able to inspect the model updates of single clients (e.g., because the updates could be aggregated via Multi-Party Computation \cite{truex2019hybrid,zhang2022augmented}). This would make unfeasible the naive solution of rolling back to the global aggregation of round $t_{u}$ \minor{and removing the individual contribution of client $u$} because it requires access to single updates of clients. \minor{Moreover, even if it could be possible to have direct access to the past model updates,} the central FL aggregator \minor{would be required} to store the entire history of aggregated models and collected updates of each round, which would translate into unfeasible storage capacity.
\\\\
\textbf{Inscrutable Local Datasets.} In centralized ML all the data used for training are supposed to be directly accessible and potentially be used for MU. On the contrary, in FL contexts, datasets remain private by design. This prevents the application of solutions from plain MU that require direct access to the data to forget and/or to the data to retain. For example, the recent work in \cite{chundawat2023can}, targeting \textit{centralized} unlearning, selectively induces noisy knowledge on the data to forget while reinforcing \textit{good knowledge} on the \textit{retain set} respectively employing an incompetent or competent teacher in a Knowledge-Distillation framework. This kind of mechanism cannot be directly adapted to FL due to the inscrutability of clients' local datasets, which cannot be exported or accessed by other parties.
\\\\
\textbf{Iterative Learning Process.} \minor{An additional factor that differentiates FU from MU is the iterative nature of FL. FL algorithms proceed in cyclical rounds, }and the global model $w_t$ aggregated at round $t$ is the starting state for round $t+1$. Suppose that at round $t_{i}$ a client $u$, which has been included in a round $t_u$ ($t_u<t_i$), requires its contribution deletion (e.g., client unlearning). The naive solution to client unlearning would be to recover the global model at round $t_u$ and to exclude client $u$'s updates from the aggregation of round $t_u$. However, the iterative nature of FL algorithms implies that all the global aggregations subsequent to round $t_u$ should be invalidated.
\\\\
\textbf{Data Heterogeneity.} In centralized ML the data is supposed to be identically and independently distributed (IID) since data are centralized in the same location and can be shuffled and arbitrarily partitioned among machines in case of distributed training. On the contrary, one of the defining characteristics of the FL setting is represented by the presence of data heterogeneity among clients, i.e., clients hold data that are likely to be non-IID. While this may hamper the convergence of the global model \cite{hsu2019measuring, li2022federated}, this trait of FL settings could be convenient for an FU algorithm, since it may be easier to erase the learned representations peculiar to a skewed subset of data (see also the illustrative results in Sec.  \ref{subsec:why_unl}). However, since data are not directly accessible, it may not be straightforward to assess the level of data heterogeneity in advance.
\\\\
\subsection{Entity Performing Unlearning} 
The unlearning phase can be performed by one or more parties in the federation, with relative benefits and drawbacks \cite{10148937}:
\begin{itemize}
    \item \textbf{Server.} The server has more computation power and storage capacity than clients. It could store the historical global models as well as keep track of historical individual clients' updates. In this way, the server could remove the contribution of a specific client, when requested to. However, to perform such a removal, the server should be able to associate one or more specific historical updates with the identity of a specific client, which poses additional privacy concerns. 
    \item \textbf{Target Client.} The target client, i.e., client $u$, has direct access to the data to be forgotten. It can locally perform and assess the unlearning success. However, this poses further security concerns when considering possibly malicious clients, who may request a local unlearning to inject a backdoor in the model or to hamper the training on purpose, taking advantage of their role in the unlearning phase. Furthermore, regular MU methods often require both retaining and forgetting data, e.g., to alternate forgetting and retaining phases so that the model can selectively unlearn. The technical issue in only-local FU is to be able to retain  \textit{good knowledge} without having access to other clients' data.   
    \item \textbf{Remaining Clients.} The remaining clients, i.e., the federation without client $u$, hold the \textit{retain data} and are typically used to recover the performance of an unlearned model.
\end{itemize}
Given the above benefits and drawbacks, mixed solutions may also be possible, e.g., target client $u$ and the server may cooperate during the unlearning phase.

\section{Design and Implementation Guidelines for Efficient FU Algorithms}\label{sec:desgud}
Here, we present the primary requirements that an FU algorithm should meet to be efficiently adopted and integrated into practical scenarios. In addition, we delve into the in-depth exploration of the metrics that have been proposed to assess the extent to which data has been effectively unlearned from the global model.

\subsection{Requirements of Federated Unlearning Algorithms}
\label{sec:requirements}
\begin{enumerate}
\item \textbf{Improved Efficiency wrt. the Retraining Strategy.} An unlearning algorithm should be more efficient than the naive, possibly unfeasible, retraining strategy, e.g., re-utilizing the knowledge that would have been discarded by a global model rollback

\item \textbf{Retained or Recovered Performances.} Either just after the unlearning process (retained performances) or after a set of new FL rounds (recovered performance), the unlearned model should attain similar performance to the original model. In particular, it should exhibit the following properties.
\begin{itemize}
    \item \textit{Comparable performances on test data wrt. the retrained model.} The performance (e.g., loss and accuracy) of the unlearned model on a test set should be comparable to the performance of a model trained only on a sanitized dataset. Since the unlearned global model needs a certain amount of rounds to recover the performance of the original model.
    \item \textit{\minor{Performance results on test data wrt. the original model.}} \minor{The performance of the unlearned model on a test set can be worse or even better than that of the original model depending on the impact of the client $u$'s data. For example, when data among clients are highly heterogeneous and a client holds a prevalent fraction or even all of the samples belonging to a class, the related performance results will likely deteriorate due to the difficulty of generalization for that class. Therefore, comparing the performance of the unlearned model with the original model may not be always informative.}
\end{itemize}

\item \textbf{Effective Forgetting.}
\begin{itemize}
    \item \textit{Comparable performances on forgetting data wrt. the retrained model.} Ideally, the prediction of an unlearned model on forgetting data should be indistinguishable from the prediction of a model trained only on the retained dataset. Referring to Table \ref{table:results_train_test}, after an ideal unlearning phase, the global model trained including client $u$ should not significantly outperform the retrained global model on client $u$'s train data.
    \item \textit{Significantly different performances on forgetting data wrt. the original model.} If the unlearning process is successful, the unlearned model and the original model should respond significantly differently when presented with forgetting data. Accordingly, to the specific metric or indicator, the unlearned model can be expected to improve or reduce the specific metric value on forgetting data (e.g., reducing a backdoor attack success rate). In general, the absolute value of the delta among the unlearned model performance and original model performance metric on forgetting data should be maximized.
\end{itemize}
\item \textbf{Same Privacy Levels as Regular FL.}  FL's main characteristic is the privacy by design that comes with leaving data distributed on the data owner's premises. An FU algorithm should not violate this fundamental requirement.
\end{enumerate}
\subsection{Metrics to Assess Unlearning}
\label{sec:metrics}

\begin{table*}[t!]
\begin{center}
{
\caption{Summary of the main metrics in FU literature.}
\label{table:metrics}
\begin{adjustbox}{width=\textwidth,center}
\begin{tabular}{l l l c c c c c c c}
\toprule
\textbf{Category} & \textbf{ID} & \textbf{Metric} & \textbf{Range} & \textbf{Exp. Value} & \textbf{Sample} & \textbf{Class} & \textbf{Client} \\
\midrule
\multirow{2}{*}{Improved Efficiency} & IE.1 & Speed-up ratio & $(0, +\infty) $ & $\uparrow, w^r$ & \checkmark & \checkmark & \checkmark \\
& IE.2 & Communication cost reduction ratio & $(0, +\infty) $ & $\uparrow, w^r$ & \checkmark & \checkmark & \checkmark \\
\midrule
\multirow{11}{*}{Recovered Performance} & RP.1 & SAPE & [0, 1] & $\downarrow, w^r$ & \checkmark & \textbf{-} & \checkmark\\
& RP.2 & Performance on test dataset & & $\approx, w^r, w^i$ & \checkmark & \textbf{-} & \checkmark\\
& & Accuracy & [0, 1] & & & & \\
& & Loss & $[0, +\infty)$ & & & & \\
& RP.3 & Performance on retained training dataset & & $\approx, w^r, w^i$ & \checkmark & \textbf{-} & \checkmark\\
& & Accuracy & [0, 1] & & & & \\
& & Loss & $[0, +\infty)$ & & & & \\
& RP.4 & Performance on retained or unlearned classes (test dataset) & & $\approx, w^r, w^i$ & \textbf{-}& \checkmark & \textbf{-} \\
& & Accuracy & [0, 1] & & & \\
& & Loss & $[0, +\infty)$ & & & & \\
& RP.5 & ECE \cite{naeini2015obtaining} & [0, 1] & $\approx, w^r, w^i$ & \textbf{-} & \textbf{-} & \checkmark \\

 \midrule
\multirow{17}{*}{Forgetting Verification} & FV.1 & Performance on client $u$'s data & & $(\approx, w^r), (\downarrow/\uparrow, w^i)$ & \checkmark & \textbf{-} & \checkmark\\
& & Accuracy & [0, 1] & & & &  \\
& & Loss & $[0, +\infty)$ & & & & \\
& FV.2 & Accuracy on removed class(es) (test dataset) & [0, 1] & $\downarrow$ & \textbf{-} & \checkmark & \textbf{-} \\
& FV.3 & KL divergence (model's output vs. uniform distrib.) & $[0, +\infty)$ & $\downarrow$ & \checkmark & \textbf{-} & \checkmark\\ 
& FV.4 & KL divergence (per-class accuracy distributions) & $[0, +\infty)$ & $\downarrow$ & \textbf{-} & \checkmark & \textbf{-}\\
& FV.5 & KL divergence between retrained and unlearned weights  & $[0, +\infty)$ & 0 & \textbf{-} & \textbf{-} & \checkmark \\
& FV.6 & $\ell_2$-norm between retrained and unlearned weights & $[0, +\infty)$ & $\downarrow$ & \checkmark & \textbf{-} & \checkmark \\
& FV.7 & FR (Liu et al. \cite{liu2020learn}) & [0, 1] & $\uparrow$ & \checkmark & \textbf{-} & \checkmark \\
 & FV.8 & IF \cite{koh2017understanding} on forgetting data & $(-\infty, +\infty)$ & 0 & \textbf{-} & \textbf{-} & \checkmark\\
 & FV.9 & RE \cite{shaik2023framu} & $[0, +\infty)$ & $\downarrow$ & \checkmark & \textbf{-} & \textbf{-}\\
 & FV.10 & AD \cite{shaik2023framu} & $[0, +\infty)$ & $(\downarrow, w^r)$, $(\uparrow, w^i)$ & \checkmark & \textbf{-} & \checkmark \\
 & FV.11 &  BASR (\cite{gu2017badnets, bagdasaryan2020backdoor, fang2020local}) & [0, 1] & $(\approx, w^r)$, $(\downarrow, w^i)$ & \checkmark & \textbf{-}  & \checkmark \\
 & FV.12 & MIA (Shokri et al. \cite{shokri2017membership}) & [0, 1] & $(\approx, w^r)$, $(\downarrow, w^i)$ & \checkmark & \textbf{-}  & \checkmark \\
 & FV.13 & MIA (Golatkar et al. \cite{golatkar2021mixed}) & [0, 1] & $(\approx, w^r)$, $(\downarrow, w^i)$ & \checkmark & \textbf{-}& \checkmark \\
 & FV.14 & MIA (Yeom et al. \cite{yeom2018privacy}) & [0, 1] & $(\approx, w^r)$, $(\downarrow, w^i)$ & \checkmark & \textbf{-}& \checkmark \\
 & FV.15 & MIB (Hu et al. \cite{hu2022membership}) & [0, 1] & $(\approx, w^r)$, $(\downarrow, w^i)$ & \checkmark & \textbf{-}& \checkmark \\

\bottomrule
\end{tabular}
\end{adjustbox}
}
\end{center}
\end{table*}


The requirements described in Subsection \ref{sec:requirements} call for metrics that can assess the efficacy of unlearning algorithms. In this subsection, we gather and introduce the metrics used in literature to evaluate FU methods and we classify them according to the scope of their assessment. Table \ref{table:metrics} summarizes the main characteristics of the considered metrics. All the reported metrics are supposed to be applied to the unlearned model $w^{\bar{u}}$. The first column indicates the scope of the metric. The \textit{ID} column associates an identifier to the metrics that will be useful in the next sections. \textit{Range} column reports the limit values of the metric. The \textit{Expected Value} column indicates if higher values or smaller values are better. If the \textit{Expected Value} column contains tuples $(\cdot, \cdot)$ in the form of $(\{\uparrow, \downarrow, \approx\}, w^*)$, it indicates if higher values, smaller values or values similar to a baseline $w^*$ (e.g., the retrained model's performances) are better. $w^i$ represents the global model before unlearning, $w^r$ represents the retrained model. Accuracy and success rate are expressed as probability and can be equivalently expressed in percentage (i.e., in a range [0\%, 100\%]). The three rightmost columns report the unlearning objectives for which the metric is most appropriate.
\\\\
\textbf{Improved Efficiency Metrics and Indicators.} This class of metrics relates to the Requirement (1) of the previous section and usually compares, as a primary dimension, the time needed by the proposed unlearning algorithm to recover performance similar to a baseline, e.g., the original model or the retrained model. For example, the ratio among $T_{w^*} / T_{w_{\bar{u}}}$, with $T_{w^*}$ being the time needed by a model trained with a specific baseline, e.g., $w^r$, and with $T_{w^{\bar{u}}}$ being the time needed by the unlearned model to attain a certain test performance, can be used as an indicator of speed-up. However, recovery time is not the only measure used in the FU literature. In \cite{halimi2022federated}, Halimi et al. consider the required cumulative communication cost in client-to-server model updates to achieve similar performance (e.g., in test accuracy) to assess the efficiency of the unlearned method compared to the retrained model, with the lower the better. This comparison could also be expressed as a ratio $Bytes_{w^*} / Bytes_{w_u}$ (higher is better), with $Bytes$ representing the size (e.g., in MB) of the cumulative client-to-server updates.
In \cite{10229075}, Su et al. also focus on minimizing the number of clients that participate in the recovery process.
\\\\
\noindent\textbf{Retained or Recovered Performance Metrics.} This class of metrics relates to the Requirement (2) of the previous section.
The straightforward metrics that can be employed to \minor{assess} whether the unlearned model achieves performances similar to a baseline model are the test performances (e.g., loss, accuracy). Another potential metric to consider is the Expected Calibration Error (ECE) \cite{naeini2015obtaining}. It is used to evaluate the calibration of probabilistic models, focusing on the alignment between predicted probabilities and actual observed frequencies. It involves dividing predictions into probability bins, comparing average predicted probabilities with observed accuracy in each bin, and calculating the average absolute differences. ECE provides a measure of how well a model's predicted probabilities reflect the true likelihood of events, indicating the model's reliability and confidence calibration. 
In \cite{liu2022right}, Liu et al. propose to use a compact indicator for comparing the test performances of the unlearned model to a reference model. Such a metric is the Symmetric Absolute Percentage Error (SAPE), computed as 
\begin{equation}
\frac{|Acc({w^u}, D_{test})-Acc({w^*}, D_{test})|}{|Acc({w^u}, D_{test})|+|Acc({w^*}, D_{test})|},    
\end{equation}
with the baseline model indicated with $w^*$ (e.g., the retrained model or the original model) and $Acc({\cdot}, D_{test})$ being the accuracy on test data. SAPE spans in a range $\in [0, 1]$, and the lower the value, the closer the unlearned model is to \minor{matching} the baseline model's test performance. To assess Class Unlearning, the test performance should be measured and compared only on the subset of test data belonging to the retained classes \cite{wang2023classdiscriminative}. 
\\\\
\textbf{Forgetting Verification Metrics and Indicators.} This class of metrics relates to the Requirement (3) of the previous section.
Comparing the performance (e.g., loss and accuracy) of the unlearned model to a baseline model (e.g., the retrained model) on client $u$'s train data can represent a preliminary indicator to \minor{assess} selective forgetting. For example, the retrained model and the unlearned model should exhibit similar performance on client $u$'s train data. However, from a real-world, practical perspective, this can be only evaluated locally by clients, since their data are private. 

Another metric used more often to evaluate FU algorithms is the success rate of backdoor attacks (BASR) \cite{gu2017badnets, bagdasaryan2020backdoor, fang2020local}. In a nutshell, the attacker client crafts malicious inputs to instill a backdoor into the model, meaning that when the model encounters a similar input, it will misclassify it. 
For example, to inject a backdoor pattern in a visual classification task, the malicious client modifies some pixels in the original images and associates wrong labels to those inputs. The success rate of the attack is the probability that the model predicts induced wrong class for any malicious inputs. From the metric perspective, backdoor attacks should induce the global model to mispredict specific inputs that embed the backdoor pattern, while the global model's performance should remain unaffected by regular inputs. The unlearning algorithm removes the contributions in the global model of such harmful backdoored training data. Hence, a successfully unlearned global model should reduce the success rate of backdoor attacks when triggered by backdoored samples but should perform well on the test dataset (i.e., similarly to the retrained model or the original model).

Inspired by \cite{gu2017badnets}, Hu et al. \cite{hu2022membership} proposed a novel membership inference technique namely Membership Inference Backdoor (MIB). In MIB, a data owner marks their samples before releasing them.  If the marked samples have been used by an unauthorized ML model, a backdoor will be injected into the model. To prove that their data was used for training, data owners can demonstrate the presence of the backdoor in the model. In the context of FU, MIB can be leveraged to evaluate unlearning methods by introducing backdoor triggers to the erased data samples and validating the unlearned model's susceptibility to backdoor attacks. As backdoored samples are hard to remove, the efficacy of unlearning these data will be even higher when removing regular data. 

Another alternative attack for assessing the effectiveness of unlearning methods is the Membership Inference Attack (MIA) \cite{shokri2017membership,yeom2018privacy,golatkar2021mixed}. MIA tries to guess whether a specific sample was included in the training set. Then, the success of MIA can be measured via standard accuracy, precision, and recall metrics, with the unlearned model and the retrained model exhibiting similar sensitivity to MIA. The rationale when employing MIA in assessing the quality of unlearning is that a model has forgotten only if an attacker cannot guess with reliable confidence the membership of the forgetting data. The pioneering MIA proposed by Shokri et al. \cite{shokri2017membership} firstly trains the so-called attack model that, once trained, will take as input the predictions of a model on client $u$'s train data and will output a membership probability, building on the assumption that if a model is very confident in his predictions then the input example is likely to be used for training. It is worth noting that membership attacks like the one proposed by Shokri et al. \cite{shokri2017membership} require that the attacker knows the type and architecture of the ML model and has access to some data from the same underlying distribution. In \cite{golatkar2021mixed}, Golatkar et al. propose a black-box MIA that trains the attacker models on the entropy of the model's output probabilities. During the training phase, the attacker model learns to predict membership by seeing samples from the training data as members and samples from the test set as non-members. In \cite{yeom2018privacy}, Yoem et al. propose an MIA that requires knowing the average training loss of the model as well as having access to its output logits. Then, a data sample is then inferred as a member if the training loss of the model on the input data point is smaller than the known average training loss. 

To assess client unlearning within a visual classification task, Halimi et al. \cite{halimi2022federated} introduce in the federation a subset of clients that hold flipped images. A successful unlearning algorithm should reduce the accuracy on the flipped samples while retaining good accuracy on regular samples\footnote{When the flip-based evaluation method presented in \cite{halimi2022federated} is used, no data augmentation is applied during training.}.

In \cite{liu2020learn}, Liu et al. introduce the Forgetting Rate (FR), a measure to evaluate the performance of unlearning. 
To calculate FR, it is necessary to use a membership oracle that can discern whether a given sample is categorized as \textit{known} or not. Given a trained ML model and an arbitrary sample $x$, an ideal membership oracle outputs \textit{true} if $x$ belongs to the training set; otherwise, outputs \textit{false}. FR is an indicator to measure how many samples are changed from the memorized set to the unknown set after unlearning, and it is calculated as the transformation rate between \textit{known} and \textit{unknown} states, i.e.:
\begin{equation}
    FR = \frac{BT}{BT + BF} \times \frac{AF}{AT + AF},
\end{equation}
denoting as BT (the number of eliminated training samples that are predicted to be \textit{true} by the oracle before unlearning), BF (the number of eliminated training samples that are predicted to be \textit{false} by the oracle before unlearning), AF (the number of eliminated training samples that are predicted to be \textit{false} by the oracle after unlearning), AT (the number of eliminated training samples that are predicted to be \textit{true} by the oracle after unlearning).

Reconstruction Error (RE) and Activation Distance (AD) are two metrics utilized in \cite{shaik2023framu}. The RE metric evaluates the model's ability to reconstruct data subjected to unlearning, with a lower score indicating enhanced performance. Simultaneously, the AD metric evaluates the indistinguishability between model outputs by measuring the average distance between the model's predictions before and after unlearning, employing the $2$-distance metric on forgetting data, the higher the better. A similar metric is used in \cite{9521274} and referred to as \textit{prediction difference}.

In \cite{gao2022verifi}, Gao et al. propose to employ metrics on specific subsets of client $u$'s forgetting data, called \textit{markers}, on which calculating more reliable indicators about unlearning performance. Gao et al., most notably, consider as markers the \textit{forgettable samples}, i.e., samples with the highest loss variance across several rounds, and \textit{erroneous samples}, i.e., the high loss samples. To assess unlearning, Gao et al. mainly track four metrics: loss, accuracy, and influence function (IF) \cite{koh2017understanding}, which estimates the impact of a training sample on model prediction, and a metric based on KL divergence. The latter is computed as follows:
\begin{equation}
\label{eq:kl_div_verifi}
    KL(f_w(S_u), p).
\end{equation}
The formulation above measures the distributional distance between a model's output on forgetting data $f_w(S_u)$ and a uniformly distributed output probability $p$, with $p = (\frac{1}{L}, ... , \frac{1}{L})$, $L$ the number of total classes, and $KL$ the KL divergence.
If the KL divergence is low or even 0, the model is not able to produce any meaningful predictions on forgetting data, i.e., the model has forgotten such data. The KL divergence can be also used to measure the distance between weight distributions, i.e., to assess the disparity between the weight distribution obtained through the unlearning process and that obtained through retraining \cite{9593225}, with lower values expected.

While the metrics detailed above are mainly employed to assess the forgetting in client or sample unlearning methods, other metrics tailored to evaluate class unlearning exist. For example, suppose a test set is divided into two subsets, i.e., the set of samples labeled with the unlearned class and the set of all the other samples. The unlearned model's accuracy on the former set should ideally be 0, whereas the accuracy on the latter set should be similar to the retrained model \cite{zhao2023momentumdegra,wang2023classdiscriminative}. However, it should be ensured that the unlearned model does not just misclassify the removed class with another. A possible metric to assess that the unlearned model spans the removed class's prediction across the remaining classes is presented in \cite{wang2023classdiscriminative}. Wang et al. \cite{wang2023classdiscriminative} \minor{measure} the KL divergence among the distribution of the unlearned model's per-class accuracy and the distribution of retrained model's per-class accuracy. The closer to 0 the distance, the more similar the distributions are.



\section{Literature Review of Federated Unlearning Methods}
\label{sec:fu_literature}
This section offers a comprehensive survey of the existing literature on FU methods. We categorize the referenced works based on the unlearning objectives, techniques, and metrics employed for evaluation. The key features of the reviewed papers are summarized in Table \ref{table:summary}. Each referenced paper is linked with the primary unlearning objective specified by the respective authors. The \textit{Where} column indicates the entities involved in the unlearning process, while the \textit{Metrics} column reports the identifiers of the metrics (refer to Table \ref{table:metrics}) employed for evaluating the effectiveness of the unlearning scheme. Additionally, the \textit{Proxy Data} column denotes whether supplementary data, typically required by the server for recovery, are necessary. Finally, we highlight whether the authors have made their solution accessible to the community (or if they plan to do so).

\begin{table*}[h]
\centering
\caption{Summary of FU referenced works.}
\label{table:summary}
\begin{adjustbox}{max width=\textwidth}
\begin{tabular}{l c c c c c c c c}
\toprule
\textbf{Objective} & \textbf{Technique} & 
\textbf{Ref.} & 
\textbf{Where} & \textbf{Metrics} & \textbf{Proxy Data} & \textbf{Open Source}\\
\midrule
\multirow{20}{*}{Client} & \multirow{5}{*}{\makecell{Re-calibration of\\Historical Updates}} & \cite{9521274} & Remaining Clients \& Server & IE.1, RP.2, FV.1, FV.12 & \textbf{-} & \checkmark 
\\
& & \cite{yuan2023federated} & Remaining Clients \& Server & RP.2, FV.11 & \textbf{-} & \textbf{-} \\ 
& & \cite{10179336} & Remaining Clients \& Server & IE.2, RP.2, FV.11 & \textbf{-} & \textbf{-} \\ 
& & \cite{gao2022verifi} & Target Client \& Server & IE.1, IE.2, FV.1, FV.3, FV.8 & \textbf{-} &\textbf{-} \\
& & \cite{guo2023FAST} & Server & IE.1, IE.2, RP.2, FV.1  & Yes & \textbf{-}\\
\cmidrule{2-7}
& \multirow{3}{*}{Knowledge Distillation} & \cite{wu2022unlearningdistillation} & Server & FV.1, FV.11 & Yes & \textbf{-}\\
& & \cite{ye2023heterogeneous} & Target Client \& Neighborhood & RP.2 & Yes & \textbf{-}\\
& & \cite{zhao2023momentumdegra} & All Clients \& Server & IE.1, RP.3, FV.1, FV.11 & \textbf{-} & \textbf{-} \\
\cmidrule{2-7}
& \multirow{3}{*}{Gradient Modification} &  \cite{halimi2022federated} & All Clients \& Server & IE.1, IE.2, RP.3, FV.11, FV.12, FV.14 & \textbf{-}  & \textbf{-} \\
& & \cite{li2023subspace} & All Clients \& Server & RP.2, FV.11 & \textbf{-}  & \textbf{-} \\
& & \cite{alam2023get} & Target Client & RP.2, FV.11 & \textbf{-}  & \checkmark \\
\cmidrule{2-7}
& \multirow{1}{*}{Clustering} & \cite{10229075} & Clients in the Cluster \& Server & IE.1, RP.2 & \textbf{-}  & \checkmark 
\\

\cmidrule{2-7}
& \multirow{4}{*}{Bayesian FL} & \cite{9593225} & Target Client & FV.5 & \textbf{-}  &  \textbf{-}
\\
& &  \cite{9820602}  & Target Client & RP.3, FV.1 & \textbf{-}  &  \textbf{-}
\\
& &  \cite{9956829}  & Target Client & RP.2, RP.3, RP.5, FV.1 & \textbf{-}  &  \textbf{-}
\\
& & \cite{wang2023bfu} & Target Client
& IE.1, RP.2, RP.3, FV.1, FV.5, FV.6, FV.15 & \textbf{-}  &  \checkmark
\\
\cmidrule{2-7}
& \multirow{1}{*}{Differential Privacy} & \cite{10189868} & Server & IE.1, RP.2, FV.12 & \textbf{-}  &  \textbf{-}
\\
\cmidrule{2-7}
& \multirow{3}{*}{Other Approaches} & \cite{9514457} & Server & IE.1, RP.2, RP.3 & \textbf{-}  &  \textbf{-}
\\
&  & \cite{pan2022machine} & Target Client \& Server & IE.1, RP.2 & \textbf{-}  &  \checkmark
\\
&  & \cite{tao2024communication} & All Clients \& Server & IE.1, RP.2, FV.12 & \textbf{-}  &  \checkmark
\\

\midrule
\multirow{2}{*}{Class} & \multirow{1}{*}{Pruning} & \cite{wang2023classdiscriminative} & Target Client & IE.1, RP.3, FV.1, FV.2, FV.4, FV.12 & \textbf{-}  &  \textbf{-}
\\
\cmidrule{2-7}
& \multirow{1}{*}{Knowledge Distillation}  & \cite{zhao2023momentumdegra} & All Clients \& Server & IE.1, RP.3, FV.1, FV.11 & \textbf{-} & \textbf{-} \\
\midrule
\multirow{12}{*}{Sample}
& \multirow{6}{*}{Gradient Modification} 
& \cite{liu2022right} & All Clients & IE.1, RP.1, RP.2  & \textbf{-} &  \checkmark \\
& & \cite{dhasade2023quickdrop}  & Target Clients \& Remaining Clients & IE.1, IE.2, RP.3, RP.4, FV.1, FV.13  & \textbf{-}  &  \checkmark
\\
& & \cite{10355067}  & Target Client \& Server & IE.1 & Yes  &  \textbf{-} 
\\
& & \cite{jin2023forgettable}  & Target Client & RP.2, RP.3, FV.1, FV.13 & Yes  &  \textbf{-} 
\\
& & \cite{liu2020learn}  & Target Client & RP.2, RP.3, FV.7 & \textbf{-}  &  \textbf{-} 
\\
& &  \cite{10234397} & Target Client & RP.2, FV.12 & \textbf{-}  &  \textbf{-}
\\
\cmidrule{2-7}
& \multirow{2}{*}{Quantization} & \cite{xiongexact} & Target Client \& Server & IE1, RP.1, FV.12 & \textbf{-}  & \checkmark
\\
& &  \cite{pmlr-v202-che23b} & All Clients \& Server & IE.1, RP.2, RP.3, FV.1 & \textbf{-}  &  \textbf{-}
\\
\cmidrule{2-7}
& \multirow{1}{*}{Reinforcement Learning} & 
 \cite{shaik2023framu}  & Target Client & RP.2, FV.9, FV.10 & \textbf{-}  &  \textbf{-} 
\\
\cmidrule{2-7}
& \multirow{3}{*}{Other Approaches} & \cite{zhu2023} & All Clients \& Server & IE.1, IE.2, RP.2, FV.1 & \textbf{-}  &  \checkmark
\\
& &\cite{pan2022machine} & Target Client \& Server & IE.1, RP.2 & \textbf{-}  &  \checkmark
\\
&  & \cite{tao2024communication} & All Clients \& Server & IE.1, RP.2, FV.12 & \textbf{-}  &  \checkmark
\\

\bottomrule
\end{tabular}
\end{adjustbox}
\end{table*}

\subsection{Client Unlearning}
Most methods proposed for client unlearning can be seen as fast retraining strategies, which design specific ways to retain as much knowledge as possible from the ongoing training while selectively overwriting the influence of forgetting data. 
\\\\
\textbf{Re-calibration of Historical Updates.} FedEraser \cite{9521274} is one of the first methods that has been proposed to perform client unlearning, which is achieved through a retraining that speeds up the recovering phase by leveraging historical parameter updates stored on the server side. 
Suppose that the current round is $t_i$ (when unlearning is requested) and that client $u$ participated in round $t_u$. FedEraser conducts a calibration phase only including the retained clients to iteratively sanitize the updates that have been produced after client $u$ participated instead of discarding them. FedEraser performs $t_i - t_u$ calibration rounds, and each round includes local training of calibrated updates and server-side aggregation of such updates to produce the sanitized global model to be broadcast at the next re-calibration round, until all the past updates have been recovered. In \cite{yuan2023federated}, Yuan et al. introduce FRU, an unlearning algorithm designed for federated recommendation systems based on FedEraser \cite{9521274}. To minimize storage space required, FRU suggests a user-item mixed semi-hard negative sampling component to decrease the number of item embedding updates and an importance-based update selection component to retain only crucial model updates. 

Cao et al. \cite{10179336} propose FedRecover, a method designed to recover a poisoned model using historical information. FedRecover utilizes retained updates from clients to estimate the update produced during retrain-from scratch, employing the Cauchy mean theorem and the L-BFGS algorithm \cite{L-BFGS} for an efficient approximation of the integrated Hessian matrix. To enhance the model's performance, it implements various strategies for fine-tuning the recovered model. These strategies include warm-up, periodic correction, abnormality fixing, and final tuning. In the warm-up phase, clients calculate exact updates and send them to the server to create buffers used by the L-BFGS algorithm in constructing Hessian matrices. These buffers are periodically updated during the periodic correction phase, where the server requests clients to send exact updates. These updates are also used to correct the recovered global model. When an estimated model update for a client is substantial, the server asks the client to send the exact update to mitigate the impact of potentially inaccurately estimated large model updates, constituting the abnormality fixing phase. Finally, in the last phase of final tuning, clients send exact updates to fine-tune the recovered global model, a step proven to enhance the model's performance.

Gao et al. \cite{gao2022verifi} propose a unified framework to perform unlearning and verification, named VeriFi. It is the first work that grants users the \textit{right to verify (RTV)}, in which participants could actively check the unlearning effect. VeriFi consists of three parts: (1) an unlearning module, (2) a verification module, and (3) an unlearning-verification mechanism, that chains (1) and (2) into an integrated pipeline.
Although the unlearning module can be any unlearning method, the authors also propose a more efficient one-step unlearning approach \textit{scale-to-unlearn} (S2U). The unlearning process is triggered only when a participant leaves during the later stages of the FL process when the model has already stabilized. S2U scales up the contribution of the clients staying in the federation and scales down the contribution of the leaving client. This adjustment is designed to align the global model more closely with the local models of those who remain while distancing it from the leaving one. The verification process is performed locally by the leaving client. It consists of two steps: \textit{marking} and \textit{checking}. The marking step involves utilizing marked data, known as markers, to fine-tune the local model. Subsequently, the checking step verifies the extent to which the global model has unlearned the markers. 

FU serves as a viable tool for eliminating contributions from malicious clients. Guo et al. \cite{guo2023FAST} introduce FAST, a protocol designed to remove the influence of byzantine participants on the global model. The server retains all the clients' updates, using this information to directly adjust the model parameters and derive the unlearning model. Specifically, the unlearning model is obtained by subtracting the contributions of malicious clients in each training round from the final global model. For every iteration, the server compares the accuracy of the current unlearning model with that of the previous one.  If the current unlearning model consistently outperforms its predecessor, it indicates that the historical influence of the malicious client persists, prompting the continuation of the step to delete contributions in the next round. Conversely, if the accuracy consistently surpasses that of the previous unlearning model, the unlearning operation proceeds until the maximum number of attempts to eliminate malicious clients' contributions is reached. However, their method may eventually lead to a loss in the predictive power for other data that are not required to be forgotten. To overcome this concern, the server maintains an additional small benchmark dataset for supplementary training of the unlearning model, thereby enhancing the overall accuracy.
\\\\
\textbf{Knowledge Distillation.} In \cite{wu2022unlearningdistillation}, Wu et al. propose a mechanism for FU based on Knowledge Distillation (KD), following a recent trend that sees many KD-based mechanisms designed to improve the FL process \cite{mora2022knowledge}. The solution in \cite{wu2022unlearningdistillation} leverages KD to quickly recover the global model's performance after the removal of the unlearning client's historical model updates. The unlearning algorithm proceeds in two server-side steps: (1) The historical updates of the unlearned client are removed from the current global model to produce a sanitized model; (2) The previous global model works as the teacher and the sanitized model works as the student. The sanitized global model mimics the outputs of the previous global model on proxy unlabeled data to recover the drop in performance after step (1). In \cite{wu2022unlearningdistillation}, clients do not directly have to participate in the unlearning phase. However, as a drawback, the solution proposed by Wu et al. requires (unlabeled) data on the server side, which may not be easily available for specific tasks. Furthermore, the server should maintain a full history of the updates collected from clients, as well as the server should be able to identify clients' updates to then respond to unlearning requests.

Zaho et al. \cite{zhao2023momentumdegra} present a knowledge erasure strategy called Momentum Degradation (MoDe). The proposed method decouples the FU process from training, making it applicable to any model architecture that has completed FL training without altering the training procedure. Unlearning consists of two phases: knowledge erasure and memory guidance. In the first phase, the pre-trained model parameters are adjusted to reduce discriminability for target data, aligning them with retrained model parameters. The server initializes a degradation model, sends it to clients for FL training, and uses its weights as the guiding direction for updating the pre-trained model. Following momentum degradation, the pre-trained model experiences reduced discriminability for all data points. The memory guidance phase restores performance by sending the pre-trained model to all clients and the degradation model to the target client. The target client uses the degradation model's output on its local dataset as pseudo labels, guiding the local training of the pre-trained model. This ensures the unlearned model's output closely resembles the retrained model's output. All clients contribute to aggregating the pre-trained model. The knowledge erasure and memory guidance steps are iterated multiple times.

Ye et al. \cite{ye2023heterogeneous} present a decentralized unlearning framework called HDUS, employing distilled (smaller) models to build erasable ensembles for each client. HDUS operates in a fully decentralized learning scenario, where clients exchange information without the need for a central server. In addition to its local main dataset and model, each client owns a distilled model and reference dataset shared among all clients. The client's objective is to align its distilled model as closely as possible with the main model, optimizing the former using a specific loss function and the reference dataset. Distilled models are exchanged with neighboring clients. During the inference phase, the client generates an ensemble model output by weighting the output of its main model and the outputs of all neighbors' distilled models. If client $u$ requests unlearning of its data, its neighbors simply exclude client $u$'s distilled model from the ensemble.
\\\\
\textbf{Gradient Modification.} In \cite{halimi2022federated}, Halimi et al. introduce a method to perform unlearning directly on the target client. The method employs projected gradient ascent to maximize the loss from a reference model, defined as the average of the models of the remaining clients. A $\ell_2$-norm sphere with $\delta$ radius around the reference model \minor{is used to constrain} the unbounded loss. To improve the performance of the unlearned model, the remaining clients run additional rounds of FL, achieving good performance in a few rounds.
Similarly, Li et al. \cite{li2023subspace} propose a subspace-based FU method (SFU) that leverages gradient ascent, constrained by the orthogonal space of input gradient spaces formed by other clients, to eliminate the target client’s contribution. SFU does not require storing intermediate updates on the clients or the server. The process includes three steps: (1) each client, except the target one, creates a representation matrix using a part of its local data and sends it to the server. A representation matrix for each layer in a neural network is formed by combining outputs from the forward pass of random samples through the network, with each column representing the output for a different sample. The privacy of the representation matrix is protected through a DP algorithm. Random factors are added to each client's layer representation to prevent the leakage of individual client data information. This addition does not affect the subspace search process due to the orthogonal nature of the matrix. (2) The target client executes some round of gradient ascent and sends the updated gradient to the server. (3) The server then performs Singular Value Decomposition (SVD) \cite{hoecker1996svd} on the set of representation matrices to get the set of input gradient subspace. It gets the orthogonal projection of the target gradient for this subspace and removes the contribution of the target client by updating the global model.

In \cite{alam2023get} 
Alam et al. shift the perspective and explore FU to eliminate backdoors for the benefit of adversaries. Once attackers have accomplished their intended objectives or suspect the possibility of detection, they may wish to eliminate the backdoors they previously injected and erase any trace of their presence. 
The proposed method leverages gradient ascent on the target client to forget its contribution. 
However, to face challenges like catastrophic forgetting, the generation of deviating models during the unlearning process, and the varying importance of individual parameters, the authors propose two strategies: memory preservation and dynamic penalization. The first one implies utilizing the benign dataset of the target client to prevent catastrophic forgetting. The second strategy implements a dynamic penalty mechanism designed to penalize diversion from the global model. Moreover, weights are also incorporated into the penalty term, assigning a unique weight to each parameter based on its significance. Introducing non-uniform penalties for all parameters may yield superior results.
The loss function is modified following the strategies depicted before. 
\\\\
\textbf{Clustering.} KNOT \cite{10229075} performs client clustering to speed-up retraining in asynchronous FL. Indeed, when unlearning is requested, retraining is conducted exclusively within client $u$'s cluster while the remaining clients are unaffected. 
\\\\
\textbf{Bayesian FL.} Bayesian FL combines Bayesian methods with FL, introducing a distribution of models to capture uncertainty instead of a fixed model. Gong et al. \cite{9593225} introduce the first FU method in a decentralized network within a Bayesian framework. It is based on exponential family parametrization and leverages local gossip-driven communication. When the unlearning client is scheduled, it removes its local variational parameter from the current global one, and the result is forwarded to the next client.
Gong et al. \cite{9820602} also propose a non-parametric federated Bayesian unlearning method, referred to as Forget-Stein Variational Gradient Descent (Forget-SVGD). Forget-SVGD is an extension of SVGD \cite{liu2016stein}, a particle-based approximate Bayesian inference method utilizing gradient-based deterministic updates, and its federated variant called Distributed SVGD (DSVGD) \cite{kassab2022federated}. After the conclusion of the FL process, one or more clients may request to forget their data. Unlearning a dataset from the variational posterior is formulated as the minimization of the unlearning free energy. This involves identifying a distribution that closely aligns with the current variational posterior while simultaneously maximizing the average training loss for the datasets designated for forgetting. An optimized version for rate-constrained channels is presented in \cite{9956829}. It applies quantization and sparsification across multiple particles. 

BFU \cite{wang2023bfu} is another Bayesian FU method that incorporates parameter self-sharing. The proposed scheme introduces an unlearning rate that balances the trade-off between forgetting the erased data and remembering the original global model. To limit potential performance degradation, data erasure and maintaining learning accuracy are considered two different objectives. 
\\\\
\textbf{Differential Privacy.} In Sec. \ref{sec:background_dp_cf}, we presented DP as an orthogonal tool for unlearning methods. In \cite{10189868}, Zhang et al. propose FedRecovery, an algorithm that embeds DP in their unlearning mechanism. Indeed, when client $u$'s requests unlearning, firstly the server removes its influence by eliminating all its gradient residuals, which are generated from clients' historical submissions. Then, the server injects Gaussian noise into the resulting model to make such an unlearned model statistically indistinguishable from the retrained one. The noise is calibrated according to the upper bound of the distance between the two models, which is estimated through the smoothness of client $u$'s loss functions. By leveraging DP, an observer cannot assess whether a model has been trained by $K$ or $K - 1$ clients.
\\\\
\textbf{Other Approaches.} Tao et al. \cite{tao2024communication} propose a framework for exact FU that leverages the concept of total variation (TV) stability, which measures the discrepancy between two distributions. A learning algorithm is $\rho$-TV-stable if the model produced using the entire dataset and the model produced using a subset of it differ by at most $\rho$.
The authors introduce FATS, a novel algorithm for fast retraining based on FedAvg and proved to be $\rho$-TV-stable.

RevFRF \cite{9514457} is a framework designed for federated Random Forests (RFs) that supports secure participant revocation. The authors delve into the revocation process from two distinctive perspectives: (1) ensuring the removal of targeted data from the RF, and (2) preempting the potential collusion of removed clients with the server to illicitly continue utilizing the outdated RF. To achieve the first level of revocation, the RevFRF traverses the Random Decision Trees (RDTs) within the trained RF and prunes all splits contributed by the participant seeking to be forgotten. Subsequently, the RF undergoes a reconstruction, wherein only the remaining RDTs are considered. Due to their isolation from each other, the rebuilding of one RDT does not affect the others. It is crucial to note, however, that this approach is not easily extendable to other ML models, and the worst-case scenario coincides with the reconstruction of the model from scratch. The second level of revocation involves executing additional computations to refresh the revoked splits with random values, rendering them inaccessible to any potentially colluding server.

In \cite{pan2022machine}, Pan et al. introduce a novel FU algorithm to perform federated K-means++. In a nutshell, clients maintain centroid vectors computed on their local data. Those vectors are then transmitted to the server, which uses them to obtain a centroid set. If a client wishes its data to be forgotten, it resets its vector of centroids to zero, prompting the server to re-execute clustering on the remaining clients' vectors.



\subsection{Class Unlearning} 
In Convolutional Neural Network (CNN) models, the visualization of feature maps activated by different channels reveals their varying contributions to distinct image categories in image classification. Wang et al. \cite{wang2023classdiscriminative} leverage this feature to introduce FU by selectively pruning channels with high discrimination on the target category to forget. Class discrimination is determined using the Term Frequency Inverse Document Frequency (TF-IDF), a statistical metric originally designed to determine the relevance of a word to a document within a collection of documents. In this context, each channel's output can be seen as a word, the feature maps representing different categories as documents, and TF-IDF evaluates the relevance between channels and categories. Once the discriminative channels have been pruned, a fine-tuning operation is applied to restore the model's performance. 

The MoDe method proposed by Zaho et al. \cite{zhao2023momentumdegra} is also presented as a solution for category removal, without modifying the algorithm introduced in the previous subsection.

\subsection{Sample Unlearning}

Given that sample unlearning can be viewed as a more specialized instance of client unlearning, these methods inherently adopt strategies inspired by the same principles. We identify similar categories of methods.
\\\\
\textbf{Gradient Modification.} Liu et al. \cite{liu2022right} propose a rapid retraining strategy that leverages a low-cost Hessian approximation method and applies it to the local training of clients, after forgetting data deletion. Specifically, at the core of the mechanism, a diagonal empirical Fisher Information Matrix (FIM) is used, with the update rule of the unlearned model employing first and second-order moments for Hessian momentum, similar to optimizer like Adam \cite{kingma2014adam}. However, this method may imply a significant reduction in performance when dealing with complex models.

Dhasade et al. \cite{dhasade2023quickdrop} introduce QuickDrop, an FU method that leverages Dataset Distillation (DD) to perform unlearning and recovery. DD condenses a large training dataset into a compact and smaller synthetic dataset \cite{geng2023survey}. However, \minor{generating} good-quality synthetic data requires many local update steps. The clients reuse the gradient updates computed during FL training to reduce the computational overhead. When an unlearning request comes, each client performs Stochastic Gradient Ascent (SGA)
on their local distilled dataset. While in the recovery phase, the network then executes recovery rounds, during which they also use the distilled data that are not extracted from the samples to forget. It is worth noting that the use of distilled data slightly decreases the performance. To avoid decreasing performance, in the recovery phase, the authors include a few original samples in the distilled datasets that nullify potential performance drops.

FedFilter \cite{10355067} is an edge caching scheme that uses FU to remove invalid and outdated data from the global model. The unlearning is initiated by the server that selects the content to be forgotten and generates a reverse gradient. Moreover, to maximize the unlearning while minimizing its impact on the model accuracy, the parameters are iterated through SGD. The reverse gradient is applied to the local model to erase the selected content locally. Finally, for the basic layer parameters, FU is achieved by aggregating and broadcasting model parameters.

Jin et al. \cite{jin2023forgettable} present the Forgettable Federated Linear Learning (2F2L) framework. 2F2L introduces a federated linear learning strategy that leverages a linear approximation of a Deep Neural Network (DNN) given by the first-order Taylor expansion. The first initial value for the weights to perform linear approximation is built using the server's dataset. With a reasonable amount of data on the server, this value is believed to be close to the optimal model weight based on the entire dataset. Having a linear approximation, it is possible to perform data removal performing gradient ascent using the quadratic forgetting loss function \cite{golatkar2021mixed}. However, to achieve removal it is necessary to compute Hessian matrix inversion, which is a complex task. For this reason, 2F2L leverages a numerical approximation based on the server's data.

Forsaken \cite{liu2020learn} is a memorization elimination algorithm that employs dummy gradients to stimulate the neurons of the ML model and erase the knowledge of specific data. Client $u$ possesses a trainable generator that produces gradients.
During each unlearning epoch, the generator is optimized based on an objective function aiming to reduce the distance between the current confidence vector and the confidence vector of a \textit{perfectly} forgotten sample rules. Then, the client produces a new dummy gradient and sends it to the server.

In \cite{10234397}, Xia et al. propose $FedME^2$, a FU framework for Digital Twins for Mobile Networks (DTMN). $FedME^2$ comprises two modules: MEval and MErase. The first one aims to build a memory evaluation model to detect whether the data to be forgotten is remembered and the degree of memory. This information is used by the MErase module to guide the privacy erasure process. On the other hand, the MErase module optimizes the local model leveraging the model loss, the MEval loss, and a penalty term, which constrains similarity between the global model before and after data erasure. The server then re-aggregates the local model from the client and completes the construction of the global model without privacy information.
\\\\
\textbf{Quantization.} Xiong et al. \cite{xiongexact} introduce Exact-Fun, a quantized FU algorithm designed to eliminate the influence of a subset of data from a target client within a global model. In instances where a client requests the removal of a specific portion of their data, the algorithm calculates a local model using the client's trained model up to that iteration. If the resulting quantized model matches the previously global model, the deletion of that data subset has no impact on the overall model. However, if quantization leads to a mismatch, retraining becomes necessary to effectively eliminate the influence of the unlearned data.

In \cite{pmlr-v202-che23b}, Che et al. extend their prior work on Prompt Certified Machine Unlearning (PCMU) \cite{NEURIPS2022_5771d9f2} to operate effectively in a federated setting. PCMU employs randomized gradient smoothing and quantization to concurrently execute training and unlearning operations, enhancing overall efficiency. The use of random smoothing on gradients implies that the MU model, directly trained on the entire dataset, shares identical gradients (and parameters) with the model retrained solely on the remaining data. This occurs within a specific certified radius, concerning the gradient change before and after data removals, and a certified budget for data removals. To implement PCMU in a federated setting, 
the authors propose a method that involves creating MU models on clients using the PCMU algorithm. These models are then reformulated as output functions of a Nemytskii operator, 
inducing a Frechet differentiable smooth manifold. This manifold possesses a global Lipschitz constant, that bounds the difference between two local MU models. At the server, a global gradient is computed by averaging gradients from all clients, resulting in the creation of a global MU model. The global Lipschitz property ensures that this model closely aligns with each local MU model on the clients within a distance determined by the Lipschitz constant. Consequently, the global MU model can maintain the certified radius and budget of data removals from the local MU models to a certain degree.
\\\\
\textbf{Reinforcement Learning.} 
In \cite{shaik2023framu}, Shaik et al. present FRAMU, an attention-based MU framework that leverages federated reinforcement learning. FRAMU can perform unlearning of outdated, irrelevant, and private data in both single-modality and multi-modality scenarios. It adopts a federated reinforcement learning architecture, employing local agents for real-time data collection and model updates, a central server for aggregation and global model updates using the FedAvg algorithm, and an attention layer to dynamically weigh the relevance of each data point in learning and unlearning. This layer acts as a specialized approximator, enhancing individual agents' learning capabilities by assigning attention scores to data points, indicating their relevance in local learning, and continually refining the model through interactions with the environment and feedback. If these scores fall below specific predetermined thresholds, the data is considered outdated or irrelevant and is consequently removed from the model.
\\\\
\textbf{Other Approaches.} The FATS algorithm, presented in \cite{tao2024communication}, can be also applied for sample unlearning by starting the retraining from the iteration preceding the inclusion of that specific sample. 

FL can be also employed with alternative data representation such as a Knowledge Graph (KG), a structured knowledge base of real-world entities and their relations in terms of triplets. KGs have also been adopted to advance representation learning techniques, i.e., KG embeddings combine entities and their relations into a unified semantic space \cite{9416312}. FL can be integrated with KG embeddings, giving birth to a novel paradigm namely \textit{Federated KG embedding learning} \cite{10.1145/3459637.3482252}. In this direction, Zhu et al. \cite{zhu2023} propose FedLU, an FL framework for heterogeneous KG embedding. FedLU uses an unlearning method to erase specific knowledge from local embeddings. On the one hand, local knowledge from clients is erased through an interference phase based on hard and soft confusions, which reflect the distance between scores of the triplet in the forgetting set and its negative samples, i.e., the set of triplets whose intersection with the local KG is empty. On the other hand, to limit the decrease in model performance caused by the interference is recovered via knowledge distillation providing the local client KG without the forgetting set. 


The framework introduced in \cite{pan2022machine} is also applicable when a client wishes to remove some of their data. If the data to be removed is not a centroid, no action is needed. However, if it is a centroid, the client needs to eliminate that centroid and select a new one by sampling from its local data using a specific distribution. The updated centroid vector is then sent to the server, which re-executes clustering on the new vector set, producing a new set of server centroids.
\section{Distilled Lessons Learned}\label{sec:lesson}
In this section, we summarize the main lessons learned from the reported in-depth review of state-of-the-art FU research. 

\begin{enumerate}
    \item \textbf{Motivation.} FU schemes are proposed as valuable mechanisms to preserve privacy and/or enhance security. From the privacy perspective, FU can guarantee the right to be forgotten. On the security side, it is a technique for mitigating the impact of malicious clients by eliminating their contributions, i.e., poisoned data and backdoors.
    \item \textbf{Empirical Evidences.} The contribution of a specific client $u$ has a tangible influence on the global model, which outputs significantly more accurate predictions when presented with client $u$'s samples if client $u$ has been included in the pool of clients. Also, the participation of client $u$ translates to evident spikes of enhanced accuracy (and reduced loss) on client $u$'s train data for that round. Natural forgetting, i.e., just excluding client $u$ from the federation in the hope of catastrophic forgetting to rapidly remove its contribution, is not a viable option. Indeed, empirical evidences show that the impact of client participation fades away very slowly when data is homogeneous, and remains concretely noticeable for a large number of subsequent rounds. It is worth noting that in the particular case of a large-scale federation with a low participation rate of clients and non-IID local datasets, natural forgetting can help the unlearning process.
    \item \textbf{Requirements and Metrics.} As we have detailed from the requirements in Sec. \ref{sec:requirements}, an unlearning algorithm should be more efficient than a retraining alternative, and the (fine-tuned) unlearned global model should achieve performance of generalization comparable to the original model. At the same time, with effective unlearning, the unlearned global model should not respond suspiciously confidently when presented with forgetting data. Then, how to verify those requirements? 
    
    While for \textit{improved efficiency} and \textit{recovered performance} there is quite a consensus on the metrics to be used (respectively speed-up ratio and test accuracy are most often employed), the works in literature are less aligned for \textit{forgetting verification}. MIA and backdoor success rates emerged as the most used assessment metrics, while often being arranged differently work by work. 
    
    \item \textbf{Methods.} The majority of the presented solutions center around the complete removal of a specific client's contributions, i.e., client unlearning. However, the algorithms designed for a particular FU objective may have broader applicability. Specifically, client unlearning coincides with sample unlearning when the samples marked for forgetting are exclusively provided by a singular client and encompass all of their data. Furthermore, the concept of client unlearning can be viewed as a specialized instance of class unlearning, applicable when only a single client contributes to the category targeted for erasure. Similarly, these considerations can be extended to sample unlearning, wherein the subset marked for forgetting corresponds to a class to be unlearned. 
\end{enumerate}

\section{Open Problems and Future Directions}
\label{sec:future_directions}

As outlined in previous sections, FU is a relatively novel concept, attracted growing interest over recent years. However, its rapid evolution has also led to the rise of new issues that necessitate careful consideration. This section delves into a comprehensive discussion of the most interesting challenges and underlines future research directions demanding further exploration.

\subsection{Standardization}
The evaluations performed on the proposed FU schemes underscore a notable deficiency in standardization, particularly concerning the choice of datasets for experiments and the metrics employed to evaluate the effectiveness of unlearning algorithms. The review of referenced papers highlights that the datasets and metrics adopted are often different. This lack of uniformity poses a challenge to conducting fair assessments and comparisons across different FU mechanisms. For instance, researchers may manipulate chosen datasets differently and utilize a diverse array of metrics, introducing variability that hinders meaningful comparisons.

To address this issue, it becomes crucial to establish a common ground by providing all researchers with the same ready-to-use datasets. These datasets should be sourced from a specific, agreed-upon data repository and pre-processed uniformly. Similarly, adopting standardized metrics for the evaluation process is essential. These measures can help ensure consistency and facilitate a more meaningful comparison among distinct FU proposals. 

However, the lack of any standardized datasets and metrics in MU and consequently in FU demonstrate that a significant amount of work needs to be done in this direction. While some efforts have been made in the context of FL (e.g., LEAF \cite{caldas2018leaf}, TensorFlow Federated datasets \cite{tffdatasets} and Flower datasets \cite{beutel2020flower, flowerdataset}), extending such standardization to MU and FU is an area that requires substantial attention and development to foster more reliable and comparable research outcomes.

\subsection{Incentive to Avoid Unlearning}
Existing literature on FU mainly focuses on novel schemes to guarantee the user's right to be forgotten and/or remove malicious contributions. Reviewing the referenced papers, we observed that incentivizing clients to be engaged and continue contributing is still an under-explored field since the design of efficient incentive mechanisms is a complex task. The performance of the unlearned model depends on the specific client who requests to leave. For the server, it is difficult to design incentives for heterogeneous individuals with multidimensional private information like training costs. Furthermore, the server must distinguish between high-quality leaving users and low-quality ones. It is worth noting that the use of incentives may also encourage strategic users to intentionally request revocation to be rewarded. 

In this direction, only Ding et al. \cite{10.1145/3565287.3610269} propose an incentive mechanism for FL and FU. Their method is based on a four-stage game that captures the interaction and information updates during the learning and unlearning process. The users’ multi-dimensional private information is summarized as one-dimensional guiding the incentive design. 

\subsection{Multiple Unlearning Request}
Although in real-world scenarios clients may continually request multiple, sequential unlearning requests, almost all the existing FU methods \minor{overlook} this condition \cite{dhasade2023quickdrop}.  

Given that FU aims to tackle a highly practical challenge, it becomes necessary to design and evaluate unlearning mechanisms in practical scenarios. In this regard, it is essential to examine the system's efficacy under the specific circumstance of handling multiple unlearning requests simultaneously. Therefore, FU schemes should satisfy the requirements of unlearning multiple classes, or the data of multiple clients using a single unlearning and recovery stage (if needed).

\subsection{Feature Unlearning}
The analysis of the literature showcases that unlearning solutions focus predominantly on horizontal FU and do not work in a vertical setting. Feature unlearning is unique to vertical FL (VFL) and aims to erase data features owned by a client $u$. This approach differs significantly from traditional unlearning methods, which typically focus on eliminating specific data samples. Consequently, techniques developed for other unlearning scenarios may not be directly applicable. Notably, client unlearning can be viewed as a specialized case of feature unlearning where a client $u$ possesses only the features targeted for unlearning. 

Despite the potential of feature unlearning is still an open challenge, as supported by the very limited body of research. Deng et al. \cite{electronics12143182} contribute to addressing this gap by presenting a method based on logistic regression for vertical FU. In the context of VFL, participants collaboratively share a common set of data samples while maintaining distinct sets of features \cite{feng2020multi}. The proposed unlearning algorithm subtracts the updates from the target client, contributing to the refinement of the global model. To enhance both efficiency and accuracy, constraints are imposed on intermediate updates, thereby minimizing the sum of intermediate parameters. This optimization strategy results in a reduced impact on the sum of intermediate parameters for the remaining clients when the target client's contributions are removed. To generate the unlearned model, the server must retain the intermediate parameters submitted by each client during the overall training. When a target client requests unlearning of their data, the server retrieves the intermediate updates from that client in the last round, negates these updates, and distributes them to the remaining clients. Subsequently, each client utilizes these negated updates to adjust their local model parameters.

\subsection{Security and Privacy}
To unlearn data from a model, the server or clients must perform some operations that lead to a novel version of the global model. The reviewed FU methods are based on the assumption that the server and clients trust each other and data involved in the training are correct \cite{guo2023FAST}. However, it is trivial to achieve this level of trust, especially in a highly distributed environment such as FL where participants are typically unknown \cite{10143976}. Additionally, many of the existing unlearning schemes are directly executed by clients and we cannot control whether they behave honestly or not.
For example, in \cite{wang2023classdiscriminative}, the TF-IDF used to determine which channel must be pruned to unlearn a class is directly computed by clients. Therefore, a malicious participant may intentionally provide a wrong value, causing the removal of another category. Therefore, we need to build trustworthy FU mechanisms that can address these issues.

Furthermore, most of the proposed FU schemes are based on gradient information from the target client or all clients. As proven in different works \cite{NEURIPS2021_5d44a2b0, 9735364, NEURIPS2020_c4ede56b}, a malicious server can leverage the client's model and the global model to reconstruct the original client's data. Therefore, several privacy-preserving techniques, such as DP (see Section \ref{sec:background_dp_cf}), Homomorphic Encryption (HE) \cite{10.1145/3214303}, and Multiparty Computation (MPC) \cite{10.1145/3383455.3422562}, have been proposed to aggregate client's updates while preserving their privacy. These privacy concerns demand FU methods that enable unlearning without leaking private information.

\subsection{Data Heterogeneity Among Clients}
Data heterogeneity among clients naturally arises in typical FL settings. As shown in \ref{subsec:why_unl}, in specific configurations characterized by low client participation rates and substantial heterogeneity among clients, achieving unlearning of client data may be easier. However, the influence of data heterogeneity on FU algorithms is often under-explored. Frequently, empirical evaluations of proposed algorithms assume heterogeneous data distribution among clients as well as small-scale federations. 

Moreover, the efficacy of proposed unlearning methods within FedAvg variants designed to address non-IID data \cite{li2022federated,mora2022federated} is still an unexplored dimension. Understanding how this specific class of algorithm either facilitates or hinders the unlearning process can provide valuable insights into the robustness and adaptability of unlearning techniques in different FL environments. 

\section{Conclusions}\label{sec:conclusion}
Privacy-preserving techniques have gained increasing attention in recent years. In this perspective, recent regulations have stated that individuals should have full control over their information, including its dynamic removal. This condition poses a serious challenge for ML, given that models are trained by employing data contributed by users. The right to eliminate the influence of specific training samples has given rise to the novel concept of MU, known as FU in federated scenarios.

This paper presents a comprehensive survey on FU, with a particular focus on technical design/implementation guidelines and practical usage scenarios. Firstly, we introduce the background concepts needed to understand the FU research area, by including the formalization of FU objectives and the clarifications of the key distinctions from traditional MU. To provide concrete insights into FU and its implications, we conduct a series of experiments by using various datasets. Subsequently, we offer valuable guidelines for designing and implementing FU algorithms, encompassing an in-depth analysis of the most employed FU evaluation metrics. Moreover, the paper offers a technically detailed review of the existing literature, by categorizing the proposed solutions according to a novel taxonomy based on the target FU objectives and the employed metrics. Finally, we contribute to the technical discussion in the field by identifying the most relevant open challenges and the most promising directions for future research work.


\bibliographystyle{IEEEtran}
\bibliography{bib} 

\begin{thebibliography}{100}
\providecommand{\url}[1]{#1}
\csname url@samestyle\endcsname
\providecommand{\newblock}{\relax}
\providecommand{\bibinfo}[2]{#2}
\providecommand{\BIBentrySTDinterwordspacing}{\spaceskip=0pt\relax}
\providecommand{\BIBentryALTinterwordstretchfactor}{4}
\providecommand{\BIBentryALTinterwordspacing}{\spaceskip=\fontdimen2\font plus
\BIBentryALTinterwordstretchfactor\fontdimen3\font minus \fontdimen4\font\relax}
\providecommand{\BIBforeignlanguage}[2]{{%
\expandafter\ifx\csname l@#1\endcsname\relax
\typeout{** WARNING: IEEEtran.bst: No hyphenation pattern has been}%
\typeout{** loaded for the language `#1'. Using the pattern for}%
\typeout{** the default language instead.}%
\else
\language=\csname l@#1\endcsname
\fi
#2}}
\providecommand{\BIBdecl}{\relax}
\BIBdecl

\bibitem{10.1145/2507157.2507163}
J.~McAuley and J.~Leskovec, ``{Hidden Factors and Hidden Topics: Understanding Rating Dimensions with Review Text},'' in \emph{Proceedings of the 7th ACM Conference on Recommender Systems}, ser. RecSys '13.\hskip 1em plus 0.5em minus 0.4em\relax New York, NY, USA: Association for Computing Machinery, 2013, p. 165–172.

\bibitem{10.1145/3477495.3532027}
Z.~Yi, X.~Wang, I.~Ounis, and C.~Macdonald, ``Multi-modal graph contrastive learning for micro-video recommendation,'' in \emph{Proceedings of the 45th International ACM SIGIR Conference on Research and Development in Information Retrieval}, ser. SIGIR '22.\hskip 1em plus 0.5em minus 0.4em\relax New York, NY, USA: Association for Computing Machinery, 2022, p. 1807–1811.

\bibitem{fbcambridge}
Wikipedia, ``Facebook–cambridge analytica data scandal,'' URL \url{https://en.wikipedia.org/wiki/Facebook-Cambridge_Analytica_data_scandal}, accessed on October 2023.

\bibitem{gdpr2016}
E.~Union, ``Complete guide to general data protection regulation compliance,'' URL \url{https://gdpr.eu/}, accessed on October 2023.

\bibitem{californiaccpa}
S.~of~California Department~of Justice, ``California consumer privacy act (ccpa),'' URL \url{https://oag.ca.gov/privacy/ccpa}, accessed on October 2023.

\bibitem{xu2023machine}
H.~Xu, T.~Zhu, L.~Zhang, W.~Zhou, and P.~S. Yu, ``Machine unlearning: A survey,'' \emph{ACM Computing Surveys}, vol.~56, no.~1, pp. 1--36, 2023.

\bibitem{7163042}
Y.~Cao and J.~Yang, ``{Towards Making Systems Forget with Machine Unlearning},'' in \emph{2015 IEEE Symposium on Security and Privacy}, 2015, pp. 463--480.

\bibitem{zhu2019deep}
L.~Zhu, Z.~Liu, and S.~Han, ``{Deep Leakage from Gradients},'' in \emph{Proc. of Conference on Neural Information Processing Systems}, 2019, pp. 14\,747--14\,756.

\bibitem{shokri2017membership}
R.~Shokri, M.~Stronati, C.~Song, and V.~Shmatikov, ``Membership inference attacks against machine learning models,'' in \emph{2017 IEEE Symposium on Security and Privacy (SP)}.\hskip 1em plus 0.5em minus 0.4em\relax IEEE, 2017, pp. 3--18.

\bibitem{liu2020learn}
Y.~Liu, Z.~Ma, X.~Liu, and J.~Ma, ``Learn to forget: User-level memorization elimination in federated learning,'' \emph{arXiv preprint arXiv:2003.10933}, vol.~1, 2020.

\bibitem{yang2023survey}
J.~Yang and Y.~Zhao, ``{A Survey of Federated Unlearning: A Taxonomy, Challenges and Future Directions},'' \emph{arXiv preprint arXiv:2310.19218}, 2023.

\bibitem{10148937}
F.~Wang, B.~Li, and B.~Li, ``Federated unlearning and its privacy threats,'' \emph{IEEE Network}, pp. 1--7, 2023.

\bibitem{liu2023survey}
Z.~Liu, Y.~Jiang, J.~Shen, M.~Peng, K.-Y. Lam, and X.~Yuan, ``A survey on federated unlearning: Challenges, methods, and future directions,'' \emph{arXiv preprint arXiv:2310.20448}, 2023.

\bibitem{zhang2023review}
H.~Zhang, T.~Nakamura, T.~Isohara, and K.~Sakurai, ``{A review on machine unlearning},'' \emph{SN Computer Science}, vol.~4, no.~4, p. 337, 2023.

\bibitem{qu2023learn}
Y.~Qu, X.~Yuan, M.~Ding, W.~Ni, T.~Rakotoarivelo, and D.~Smith, ``{Learn to Unlearn: A Survey on Machine Unlearning},'' \emph{arXiv preprint arXiv:2305.07512}, 2023.

\bibitem{nguyen2022survey}
T.~T. Nguyen, T.~T. Huynh, P.~L. Nguyen, A.~W.-C. Liew, H.~Yin, and Q.~V.~H. Nguyen, ``A survey of machine unlearning,'' \emph{arXiv preprint arXiv:2209.02299}, 2022.

\bibitem{golatkar2020eternal}
A.~Golatkar, A.~Achille, and S.~Soatto, ``Eternal sunshine of the spotless net: Selective forgetting in deep networks,'' in \emph{Proceedings of the IEEE/CVF Conference on Computer Vision and Pattern Recognition}, 2020, pp. 9304--9312.

\bibitem{mcmahan2017communication}
B.~McMahan, E.~Moore, D.~Ramage, S.~Hampson, and B.~A. y~Arcas, ``Communication-efficient learning of deep networks from decentralized data,'' in \emph{Artificial intelligence and statistics}.\hskip 1em plus 0.5em minus 0.4em\relax PMLR, 2017, pp. 1273--1282.

\bibitem{bellavista2021decentralised}
P.~Bellavista, L.~Foschini, and A.~Mora, ``{Decentralised Learning in Federated Deployment Environments: A System-Level Survey},'' \emph{ACM Computing Surveys (CSUR)}, vol.~54, no.~1, pp. 1--38, 2021.

\bibitem{reddi2020adaptive}
S.~Reddi, Z.~Charles, M.~Zaheer, Z.~Garrett, K.~Rush, J.~Kone{\v{c}}n{\`y}, S.~Kumar, and H.~B. McMahan, ``Adaptive federated optimization,'' \emph{arXiv preprint arXiv:2003.00295}, 2020.

\bibitem{melis2019exploiting}
L.~Melis, C.~Song, E.~De~Cristofaro, and V.~Shmatikov, ``Exploiting unintended feature leakage in collaborative learning,'' in \emph{2019 IEEE Symposium on Security and Privacy (SP)}.\hskip 1em plus 0.5em minus 0.4em\relax IEEE, 2019, pp. 691--706.

\bibitem{nasr2018comprehensive}
M.~Nasr, R.~Shokri, and A.~Houmansadr, ``Comprehensive privacy analysis of deep learning: Stand-alone and federated learning under passive and active white-box inference attacks,'' \emph{arXiv preprint arXiv:1812.00910}, 2018.

\bibitem{lesort2020continual}
T.~Lesort, V.~Lomonaco, A.~Stoian, D.~Maltoni, D.~Filliat, and N.~D{\'\i}az-Rodr{\'\i}guez, ``Continual learning for robotics: Definition, framework, learning strategies, opportunities and challenges,'' \emph{Information fusion}, vol.~58, pp. 52--68, 2020.

\bibitem{kemker2018measuring}
R.~Kemker, M.~McClure, A.~Abitino, T.~Hayes, and C.~Kanan, ``Measuring catastrophic forgetting in neural networks,'' in \emph{Proceedings of the AAAI conference on artificial intelligence}, vol.~32, no.~1, 2018.

\bibitem{legate2023re}
G.~Legate, L.~Caccia, and E.~Belilovsky, ``Re-weighted softmax cross-entropy to control forgetting in federated learning,'' \emph{arXiv preprint arXiv:2304.05260}, 2023.

\bibitem{caldarola2022improving}
D.~Caldarola, B.~Caputo, and M.~Ciccone, ``{Improving Generalization in Federated Learning by Seeking Flat Minima},'' in \emph{Proc. of European Computer Vision Conference}.\hskip 1em plus 0.5em minus 0.4em\relax Springer, 2022, pp. 654--672.

\bibitem{lee2021preservation}
G.~Lee, M.~Jeong, Y.~Shin, S.~Bae, and S.-Y. Yun, ``Preservation of the global knowledge by not-true distillation in federated learning,'' in \emph{Advances in Neural Information Processing Systems}, 2022.

\bibitem{gao2022verifi}
X.~Gao, X.~Ma, J.~Wang, Y.~Sun, B.~Li, S.~Ji, P.~Cheng, and J.~Chen, ``Verifi: Towards verifiable federated unlearning,'' \emph{arXiv preprint arXiv:2205.12709}, 2022.

\bibitem{dwork2011differential}
C.~Dwork, ``Differential privacy,'' \emph{Encyclopedia of Cryptography and Security}, pp. 338--340, 2011.

\bibitem{dwork2014algorithmic}
C.~Dwork, A.~Roth \emph{et~al.}, ``The algorithmic foundations of differential privacy,'' \emph{Foundations and Trends{\textregistered} in Theoretical Computer Science}, vol.~9, no. 3--4, pp. 211--407, 2014.

\bibitem{mcmahan2017learning}
H.~B. McMahan, D.~Ramage, K.~Talwar, and L.~Zhang, ``Learning differentially private language models without losing accuracy,'' \emph{arXiv preprint arXiv:1710.06963}, 2017.

\bibitem{geyer2017differentially}
R.~C. Geyer, T.~Klein, and M.~Nabi, ``Differentially private federated learning: A client level perspective,'' \emph{arXiv preprint arXiv:1712.07557}, 2017.

\bibitem{bagdasaryan2019differential}
E.~Bagdasaryan, O.~Poursaeed, and V.~Shmatikov, ``Differential privacy has disparate impact on model accuracy,'' in \emph{Advances in Neural Information Processing Systems}, 2019, pp. 15\,453--15\,462.

\bibitem{wu2022federated}
L.~Wu, S.~Guo, J.~Wang, Z.~Hong, J.~Zhang, and Y.~Ding, ``Federated unlearning: Guarantee the right of clients to forget,'' \emph{IEEE Network}, vol.~36, no.~5, pp. 129--135, 2022.

\bibitem{xie2021segformer}
E.~Xie, W.~Wang, Z.~Yu, A.~Anandkumar, J.~M. Alvarez, and P.~Luo, ``Segformer: Simple and efficient design for semantic segmentation with transformers,'' \emph{Advances in Neural Information Processing Systems}, vol.~34, pp. 12\,077--12\,090, 2021.

\bibitem{truex2019hybrid}
S.~Truex, N.~Baracaldo, A.~Anwar, T.~Steinke, H.~Ludwig, R.~Zhang, and Y.~Zhou, ``{A Hybrid Approach to Privacy-preserving Federated Learning},'' in \emph{Proc. of the ACM Workshop on Artificial Intelligence and Security}, 2019, pp. 1--11.

\bibitem{zhang2022augmented}
C.~Zhang, S.~Ekanut, L.~Zhen, and Z.~Li, ``Augmented multi-party computation against gradient leakage in federated learning,'' \emph{IEEE Transactions on Big Data}, 2022.

\bibitem{chundawat2023can}
V.~S. Chundawat, A.~K. Tarun, M.~Mandal, and M.~Kankanhalli, ``Can bad teaching induce forgetting? unlearning in deep networks using an incompetent teacher,'' in \emph{Proceedings of the AAAI Conference on Artificial Intelligence}, vol.~37, no.~6, 2023, pp. 7210--7217.

\bibitem{hsu2019measuring}
T.-M.~H. Hsu, H.~Qi, and M.~Brown, ``Measuring the effects of non-identical data distribution for federated visual classification,'' \emph{arXiv preprint arXiv:1909.06335}, 2019.

\bibitem{li2022federated}
Q.~Li, Y.~Diao, Q.~Chen, and B.~He, ``{Federated Learning on Non-iid Data Silos: An Experimental Study},'' in \emph{Proc. of IEEE International Conference on Data Engineering (ICDE)}.\hskip 1em plus 0.5em minus 0.4em\relax IEEE, 2022, pp. 965--978.

\bibitem{naeini2015obtaining}
M.~P. Naeini, G.~Cooper, and M.~Hauskrecht, ``Obtaining well calibrated probabilities using bayesian binning,'' in \emph{Proceedings of the AAAI conference on artificial intelligence}, vol.~29, no.~1, 2015.

\bibitem{koh2017understanding}
P.~W. Koh and P.~Liang, ``Understanding black-box predictions via influence functions,'' in \emph{International conference on machine learning}.\hskip 1em plus 0.5em minus 0.4em\relax PMLR, 2017, pp. 1885--1894.

\bibitem{shaik2023framu}
T.~Shaik, X.~Tao, L.~Li, H.~Xie, T.~Cai, X.~Zhu, and Q.~Li, ``{FRAMU: Attention-based Machine Unlearning using Federated Reinforcement Learning},'' \emph{arXiv preprint arXiv:2309.10283}, 2023.

\bibitem{gu2017badnets}
T.~Gu, B.~Dolan-Gavitt, and S.~Garg, ``Badnets: Identifying vulnerabilities in the machine learning model supply chain,'' \emph{arXiv preprint arXiv:1708.06733}, 2017.

\bibitem{bagdasaryan2020backdoor}
E.~Bagdasaryan, A.~Veit, Y.~Hua, D.~Estrin, and V.~Shmatikov, ``How to backdoor federated learning,'' in \emph{International conference on artificial intelligence and statistics}.\hskip 1em plus 0.5em minus 0.4em\relax PMLR, 2020, pp. 2938--2948.

\bibitem{fang2020local}
M.~Fang, X.~Cao, J.~Jia, and N.~Gong, ``Local model poisoning attacks to $\{$Byzantine-Robust$\}$ federated learning,'' in \emph{29th USENIX security symposium (USENIX Security 20)}, 2020, pp. 1605--1622.

\bibitem{golatkar2021mixed}
A.~Golatkar, A.~Achille, A.~Ravichandran, M.~Polito, and S.~Soatto, ``{Mixed-privacy forgetting in deep networks},'' in \emph{Proceedings of the IEEE/CVF conference on computer vision and pattern recognition}, 2021, pp. 792--801.

\bibitem{yeom2018privacy}
S.~Yeom, I.~Giacomelli, M.~Fredrikson, and S.~Jha, ``Privacy risk in machine learning: Analyzing the connection to overfitting,'' in \emph{2018 IEEE 31st computer security foundations symposium (CSF)}.\hskip 1em plus 0.5em minus 0.4em\relax IEEE, 2018, pp. 268--282.

\bibitem{hu2022membership}
H.~Hu, Z.~Salcic, G.~Dobbie, J.~Chen, L.~Sun, and X.~Zhang, ``{Membership inference via backdooring},'' in \emph{The 31st International Joint Conference on Artificial Intelligence (IJCAI-22)}, 2022.

\bibitem{halimi2022federated}
A.~Halimi, S.~Kadhe, A.~Rawat, and N.~Baracaldo, ``Federated unlearning: How to efficiently erase a client in fl?'' \emph{arXiv preprint arXiv:2207.05521}, 2022.

\bibitem{10229075}
N.~Su and B.~Li, ``Asynchronous federated unlearning,'' in \emph{IEEE INFOCOM 2023 - IEEE Conference on Computer Communications}, 2023, pp. 1--10.

\bibitem{liu2022right}
Y.~Liu, L.~Xu, X.~Yuan, C.~Wang, and B.~Li, ``The right to be forgotten in federated learning: An efficient realization with rapid retraining,'' in \emph{IEEE INFOCOM 2022-IEEE Conference on Computer Communications}.\hskip 1em plus 0.5em minus 0.4em\relax IEEE, 2022, pp. 1749--1758.

\bibitem{wang2023classdiscriminative}
J.~Wang, S.~Guo, X.~Xie, and H.~Qi, ``{Federated Unlearning via Class-Discriminative Pruning},'' in \emph{Proceedings of the ACM Web Conference 2022}, ser. WWW '22.\hskip 1em plus 0.5em minus 0.4em\relax New York, NY, USA: Association for Computing Machinery, 2022, p. 622–632.

\bibitem{9521274}
G.~Liu, X.~Ma, Y.~Yang, C.~Wang, and J.~Liu, ``Federaser: Enabling efficient client-level data removal from federated learning models,'' in \emph{2021 IEEE/ACM 29th International Symposium on Quality of Service (IWQOS)}, 2021, pp. 1--10.

\bibitem{9593225}
J.~Gong, O.~Simeone, and J.~Kang, ``Bayesian variational federated learning and unlearning in decentralized networks,'' in \emph{2021 IEEE 22nd International Workshop on Signal Processing Advances in Wireless Communications (SPAWC)}, 2021, pp. 216--220.

\bibitem{zhao2023momentumdegra}
Y.~Zhao, P.~Wang, H.~Qi, J.~Huang, Z.~Wei, and Q.~Zhang, ``Federated unlearning with momentum degradation,'' \emph{IEEE Internet of Things Journal}, pp. 1--1, 2023.

\bibitem{yuan2023federated}
W.~Yuan, H.~Yin, F.~Wu, S.~Zhang, T.~He, and H.~Wang, ``Federated unlearning for on-device recommendation,'' in \emph{Proceedings of the Sixteenth ACM International Conference on Web Search and Data Mining}, 2023, pp. 393--401.

\bibitem{10179336}
X.~Cao, J.~Jia, Z.~Zhang, and N.~Z. Gong, ``Fedrecover: Recovering from poisoning attacks in federated learning using historical information,'' in \emph{2023 IEEE Symposium on Security and Privacy (SP)}, 2023, pp. 1366--1383.

\bibitem{guo2023FAST}
X.~Guo, P.~Wang, S.~Qiu, W.~Song, Q.~Zhang, X.~Wei, and D.~Zhou, ``{FAST: Adopting Federated Unlearning to Eliminating Malicious Terminals at Server Side},'' \emph{IEEE Transactions on Network Science and Engineering}, pp. 1--14, 2023.

\bibitem{wu2022unlearningdistillation}
C.~Wu, S.~Zhu, and P.~Mitra, ``Federated unlearning with knowledge distillation,'' \emph{arXiv preprint arXiv:2201.09441}, 2022.

\bibitem{ye2023heterogeneous}
G.~Ye, Q.~V.~H. Nguyen, and H.~Yin, ``{Heterogeneous Decentralized Machine Unlearning with Seed Model Distillation},'' \emph{arXiv preprint arXiv:2308.13269}, 2023.

\bibitem{li2023subspace}
G.~Li, L.~Shen, Y.~Sun, Y.~Hu, H.~Hu, and D.~Tao, ``Subspace based federated unlearning,'' \emph{arXiv preprint arXiv:2302.12448}, 2023.

\bibitem{alam2023get}
M.~Alam, H.~Lamri, and M.~Maniatakos, ``Get rid of your trail: Remotely erasing backdoors in federated learning,'' \emph{arXiv preprint arXiv:2304.10638}, 2023.

\bibitem{9820602}
J.~Gong, J.~Kang, O.~Simeone, and R.~Kassab, ``Forget-svgd: Particle-based bayesian federated unlearning,'' in \emph{2022 IEEE Data Science and Learning Workshop (DSLW)}, 2022, pp. 1--6.

\bibitem{9956829}
J.~Gong, O.~Simeone, and J.~Kang, ``Compressed particle-based federated bayesian learning and unlearning,'' \emph{IEEE Communications Letters}, vol.~27, no.~2, pp. 556--560, 2023.

\bibitem{wang2023bfu}
W.~Wang, Z.~Tian, C.~Zhang, A.~Liu, and S.~Yu, ``{BFU: Bayesian Federated Unlearning with Parameter Self-Sharing},'' in \emph{Proceedings of the 2023 ACM Asia Conference on Computer and Communications Security}, ser. ASIA CCS '23.\hskip 1em plus 0.5em minus 0.4em\relax New York, NY, USA: Association for Computing Machinery, 2023, p. 567–578.

\bibitem{10189868}
L.~Zhang, T.~Zhu, H.~Zhang, P.~Xiong, and W.~Zhou, ``{FedRecovery: Differentially Private Machine Unlearning for Federated Learning Frameworks},'' \emph{IEEE Transactions on Information Forensics and Security}, vol.~18, pp. 4732--4746, 2023.

\bibitem{9514457}
Y.~Liu, Z.~Ma, Y.~Yang, X.~Liu, J.~Ma, and K.~Ren, ``{RevFRF: Enabling Cross-Domain Random Forest Training With Revocable Federated Learning},'' \emph{IEEE Transactions on Dependable and Secure Computing}, vol.~19, no.~6, pp. 3671--3685, 2022.

\bibitem{pan2022machine}
C.~Pan, J.~Sima, S.~Prakash, V.~Rana, and O.~Milenkovic, ``Machine unlearning of federated clusters,'' \emph{arXiv preprint arXiv:2210.16424}, 2022.

\bibitem{tao2024communication}
Y.~Tao, C.-L. Wang, M.~Pan, D.~Yu, X.~Cheng, and D.~Wang, ``Communication efficient and provable federated unlearning,'' \emph{arXiv preprint arXiv:2401.11018}, 2024.

\bibitem{dhasade2023quickdrop}
A.~Dhasade, Y.~Ding, S.~Guo, A.-m. Kermarrec, M.~De~Vos, and L.~Wu, ``{QuickDrop: Efficient Federated Unlearning by Integrated Dataset Distillation},'' \emph{arXiv preprint arXiv:2311.15603}, 2023.

\bibitem{10355067}
P.~Wang, Z.~Yan, M.~S. Obaidat, Z.~Yuan, L.~Yang, J.~Zhang, Z.~Wei, and Q.~Zhang, ``{Edge Caching with Federated Unlearning for Low-latency V2X Communications},'' \emph{IEEE Communications Magazine}, pp. 1--7, 2023.

\bibitem{jin2023forgettable}
R.~Jin, M.~Chen, Q.~Zhang, and X.~Li, ``{Forgettable Federated Linear Learning with Certified Data Removal},'' \emph{arXiv preprint arXiv:2306.02216}, 2023.

\bibitem{10234397}
H.~Xia, S.~Xu, J.~Pei, R.~Zhang, Z.~Yu, W.~Zou, L.~Wang, and C.~Liu, ``{FedME2: Memory Evaluation \& Erase Promoting Federated Unlearning in DTMN},'' \emph{IEEE Journal on Selected Areas in Communications}, vol.~41, no.~11, pp. 3573--3588, 2023.

\bibitem{xiongexact}
Z.~Xiong, W.~Li, Y.~Li, and Z.~Cai, ``{Exact-Fun: An Exact and Efficient Federated Unlearning Approach}.''

\bibitem{pmlr-v202-che23b}
T.~Che, Y.~Zhou, Z.~Zhang, L.~Lyu, J.~Liu, D.~Yan, D.~Dou, and J.~Huan, ``Fast federated machine unlearning with nonlinear functional theory,'' in \emph{Proceedings of the 40th International Conference on Machine Learning}, ser. Proceedings of Machine Learning Research, A.~Krause, E.~Brunskill, K.~Cho, B.~Engelhardt, S.~Sabato, and J.~Scarlett, Eds., vol. 202.\hskip 1em plus 0.5em minus 0.4em\relax PMLR, 23--29 Jul 2023, pp. 4241--4268.

\bibitem{zhu2023}
X.~Zhu, G.~Li, and W.~Hu, ``{Heterogeneous Federated Knowledge Graph Embedding Learning and Unlearning},'' in \emph{Proceedings of the ACM Web Conference 2023}, ser. WWW '23.\hskip 1em plus 0.5em minus 0.4em\relax New York, NY, USA: Association for Computing Machinery, 2023, p. 2444–2454.

\bibitem{L-BFGS}
J.~Nocedal, ``Updating quasi-newton matrices with limited storage,'' \emph{Mathematics of Computation}, vol.~35, no. 151, pp. 773--782, 1980.

\bibitem{mora2022knowledge}
A.~Mora, I.~Tenison, P.~Bellavista, and I.~Rish, ``Knowledge distillation for federated learning: a practical guide,'' \emph{arXiv preprint arXiv:2211.04742}, 2022.

\bibitem{hoecker1996svd}
A.~Hoecker and V.~Kartvelishvili, ``Svd approach to data unfolding,'' \emph{Nuclear Instruments and Methods in Physics Research Section A: Accelerators, Spectrometers, Detectors and Associated Equipment}, vol. 372, no.~3, pp. 469--481, 1996.

\bibitem{liu2016stein}
Q.~Liu and D.~Wang, ``Stein variational gradient descent: A general purpose bayesian inference algorithm,'' \emph{Advances in neural information processing systems}, vol.~29, 2016.

\bibitem{kassab2022federated}
R.~Kassab and O.~Simeone, ``Federated generalized bayesian learning via distributed stein variational gradient descent,'' \emph{IEEE Transactions on Signal Processing}, vol.~70, pp. 2180--2192, 2022.

\bibitem{kingma2014adam}
D.~P. Kingma and J.~Ba, ``Adam: A method for stochastic optimization,'' \emph{arXiv preprint arXiv:1412.6980}, 2014.

\bibitem{geng2023survey}
J.~Geng, Z.~Chen, Y.~Wang, H.~Woisetschlaeger, S.~Schimmler, R.~Mayer, Z.~Zhao, and C.~Rong, ``{A Survey on Dataset Distillation: Approaches, Applications and Future Directions},'' \emph{arXiv preprint arXiv:2305.01975}, 2023.

\bibitem{NEURIPS2022_5771d9f2}
\BIBentryALTinterwordspacing
Z.~Zhang, Y.~Zhou, X.~Zhao, T.~Che, and L.~Lyu, ``Prompt certified machine unlearning with randomized gradient smoothing and quantization,'' in \emph{Advances in Neural Information Processing Systems}, S.~Koyejo, S.~Mohamed, A.~Agarwal, D.~Belgrave, K.~Cho, and A.~Oh, Eds., vol.~35.\hskip 1em plus 0.5em minus 0.4em\relax Curran Associates, Inc., 2022, pp. 13\,433--13\,455. [Online]. Available: \url{https://proceedings.neurips.cc/paper_files/paper/2022/file/5771d9f214b75be6ff20f63bba315644-Paper-Conference.pdf}
\BIBentrySTDinterwordspacing

\bibitem{9416312}
S.~Ji, S.~Pan, E.~Cambria, P.~Marttinen, and P.~S. Yu, ``{A Survey on Knowledge Graphs: Representation, Acquisition, and Applications},'' \emph{IEEE Transactions on Neural Networks and Learning Systems}, vol.~33, no.~2, pp. 494--514, 2022.

\bibitem{10.1145/3459637.3482252}
\BIBentryALTinterwordspacing
H.~Peng, H.~Li, Y.~Song, V.~Zheng, and J.~Li, ``{Differentially Private Federated Knowledge Graphs Embedding},'' in \emph{Proceedings of the 30th ACM International Conference on Information \& Knowledge Management}, ser. CIKM '21.\hskip 1em plus 0.5em minus 0.4em\relax New York, NY, USA: Association for Computing Machinery, 2021, p. 1416–1425. [Online]. Available: \url{https://doi.org/10.1145/3459637.3482252}
\BIBentrySTDinterwordspacing

\bibitem{caldas2018leaf}
S.~Caldas, S.~M.~K. Duddu, P.~Wu, T.~Li, J.~Kone{\v{c}}n{\`y}, H.~B. McMahan, V.~Smith, and A.~Talwalkar, ``Leaf: A benchmark for federated settings,'' \emph{arXiv preprint arXiv:1812.01097}, 2018.

\bibitem{tffdatasets}
TensorFlow, ``Tensorflow federated,'' \url{https://flower.dev/docs/datasets/ d_2}, 2023.

\bibitem{beutel2020flower}
D.~J. Beutel, T.~Topal, A.~Mathur, X.~Qiu, J.~Fernandez-Marques, Y.~Gao, L.~Sani, K.~H. Li, T.~Parcollet, P.~P.~B. de~Gusm{\~a}o \emph{et~al.}, ``Flower: A friendly federated learning research framework,'' \emph{arXiv preprint arXiv:2007.14390}, 2020.

\bibitem{flowerdataset}
Flower, ``Flower datasets,'' \url{https://www.tensorflow.org/federated}, 2023.

\bibitem{10.1145/3565287.3610269}
N.~Ding, Z.~Sun, E.~Wei, and R.~Berry, ``Incentive mechanism design for federated learning and unlearning,'' in \emph{Proceedings of the Twenty-Fourth International Symposium on Theory, Algorithmic Foundations, and Protocol Design for Mobile Networks and Mobile Computing}, ser. MobiHoc '23.\hskip 1em plus 0.5em minus 0.4em\relax New York, NY, USA: Association for Computing Machinery, 2023, p. 11–20.

\bibitem{electronics12143182}
Z.~Deng, Z.~Han, C.~Ma, M.~Ding, L.~Yuan, C.~Ge, and Z.~Liu, ``Vertical federated unlearning on the logistic regression model,'' \emph{Electronics}, vol.~12, no.~14, 2023.

\bibitem{feng2020multi}
S.~Feng and H.~Yu, ``Multi-participant multi-class vertical federated learning,'' \emph{arXiv preprint arXiv:2001.11154}, 2020.

\bibitem{10143976}
C.~Mazzocca, N.~Romandini, M.~Mendula, R.~Montanari, and P.~Bellavista, ``{TruFLaaS: Trustworthy Federated Learning as a Service},'' \emph{IEEE Internet of Things Journal}, vol.~10, no.~24, pp. 21\,266--21\,281, 2023.

\bibitem{NEURIPS2021_5d44a2b0}
K.~Singhal, H.~Sidahmed, Z.~Garrett, S.~Wu, J.~Rush, and S.~Prakash, ``{Federated Reconstruction: Partially Local Federated Learning},'' in \emph{Advances in Neural Information Processing Systems}, M.~Ranzato, A.~Beygelzimer, Y.~Dauphin, P.~Liang, and J.~W. Vaughan, Eds., vol.~34.\hskip 1em plus 0.5em minus 0.4em\relax Curran Associates, Inc., 2021, pp. 11\,220--11\,232.

\bibitem{9735364}
C.~Chen, L.~Lyu, H.~Yu, and G.~Chen, ``{Practical Attribute Reconstruction Attack Against Federated Learning},'' \emph{IEEE Transactions on Big Data}, pp. 1--1, 2022.

\bibitem{NEURIPS2020_c4ede56b}
J.~Geiping, H.~Bauermeister, H.~Dr\"{o}ge, and M.~Moeller, ``{Inverting Gradients - How easy is it to break privacy in federated learning?}'' in \emph{Advances in Neural Information Processing Systems}, H.~Larochelle, M.~Ranzato, R.~Hadsell, M.~Balcan, and H.~Lin, Eds., vol.~33.\hskip 1em plus 0.5em minus 0.4em\relax Curran Associates, Inc., 2020, pp. 16\,937--16\,947.

\bibitem{10.1145/3214303}
A.~Acar, H.~Aksu, A.~S. Uluagac, and M.~Conti, ``{A Survey on Homomorphic Encryption Schemes: Theory and Implementation},'' \emph{ACM Comput. Surv.}, vol.~51, no.~4, jul 2018.

\bibitem{10.1145/3383455.3422562}
D.~Byrd and A.~Polychroniadou, ``{Differentially Private Secure Multi-Party Computation for Federated Learning in Financial Applications},'' in \emph{Proceedings of the First ACM International Conference on AI in Finance}, ser. ICAIF '20.\hskip 1em plus 0.5em minus 0.4em\relax New York, NY, USA: Association for Computing Machinery, 2021.

\bibitem{mora2022federated}
A.~Mora, D.~Fantini, and P.~Bellavista, ``{Federated Learning Algorithms with Heterogeneous Data Distributions: An Empirical Evaluation},'' in \emph{Proc. of IEEE/ACM Symposium on Edge Computing (SEC)}.\hskip 1em plus 0.5em minus 0.4em\relax IEEE, 2022, pp. 336--341.

\bibitem{Krizhevsky09learningmultiple}
A.~Krizhevsky, ``Learning multiple layers of features from tiny images,'' Tech. Rep., 2009.

\bibitem{WelinderEtal2010}
P.~Welinder, S.~Branson, T.~Mita, C.~Wah, F.~Schroff, S.~Belongie, and P.~Perona, ``{Caltech-UCSD Birds 200},'' California Institute of Technology, Tech. Rep. CNS-TR-2010-001, 2010.

\bibitem{maji13fine-grained}
S.~Maji, J.~Kannala, E.~Rahtu, M.~Blaschko, and A.~Vedaldi, ``Fine-grained visual classification of aircraft,'' Tech. Rep., 2013.

\bibitem{hu2023federated}
E.~Hu, Y.~Tang, A.~Kyrillidis, and C.~Jermaine, ``Federated learning over images: Vertical decompositions and pre-trained backbones are difficult to beat,'' in \emph{Proceedings of the IEEE/CVF International Conference on Computer Vision}, 2023, pp. 19\,385--19\,396.

\end{thebibliography}



\vspace{4pt}

\vspace{-33pt}
\begin{IEEEbiography}[{\includegraphics[width=1in,height=1.25in,clip,keepaspectratio]{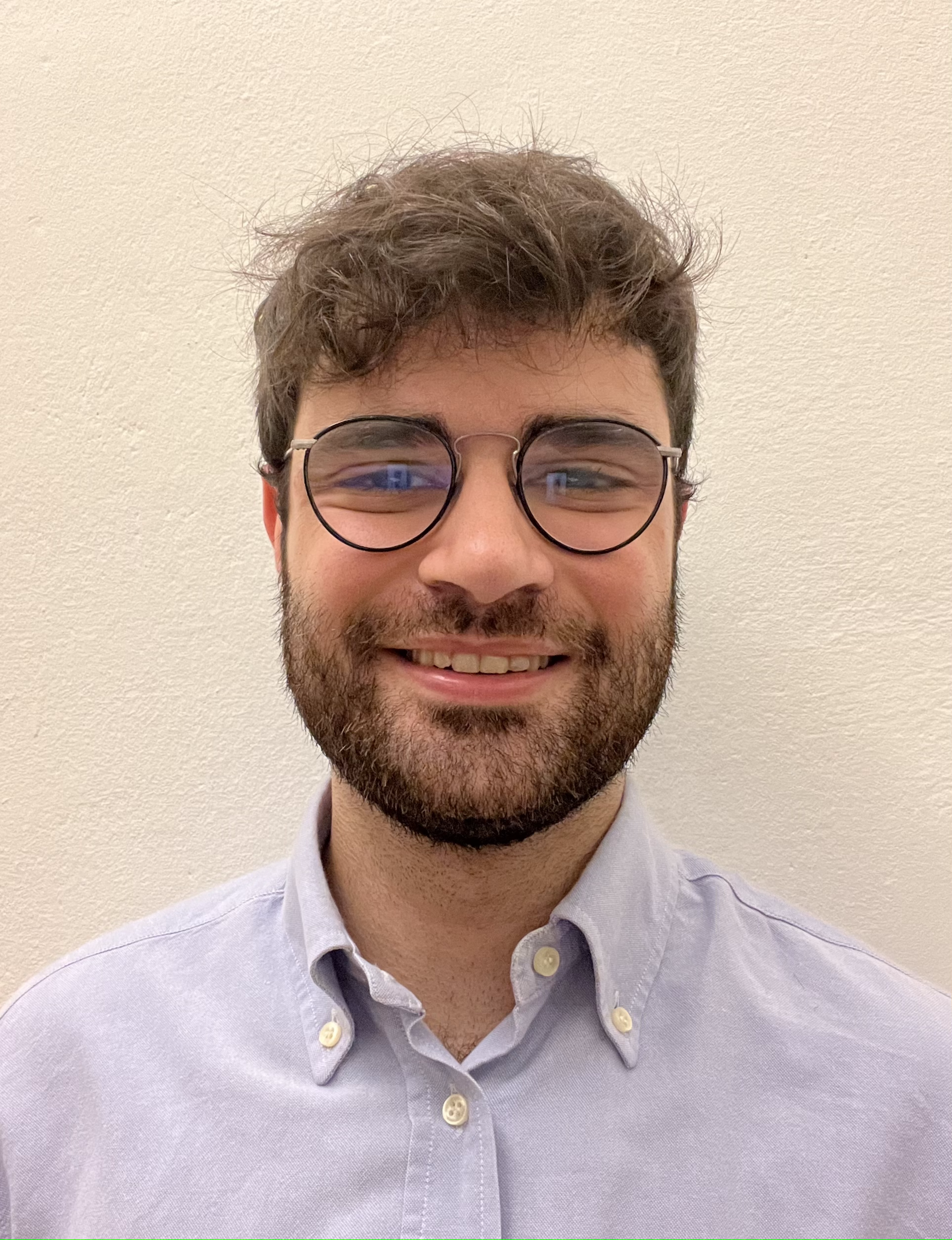}}]{Nicolò Romandini}
graduated from the University of Bologna, Italy, where he received an M.Sc. degree in computer science engineering, in 2021. He is currently a  Ph.D. student at the  Department of  Computer Science and  Engineering at the University of  Bologna. His research focuses mainly on blockchain, cybersecurity, and machine learning, and how to integrate them into IoT domains.
\end{IEEEbiography}

\vspace{-34pt}
\begin{IEEEbiography}[{\includegraphics[width=1in,height=1.25in,clip,keepaspectratio]{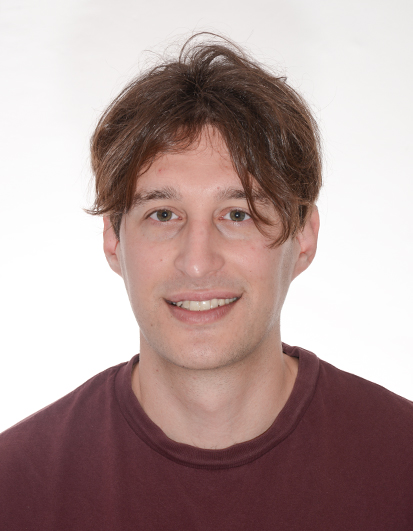}}]{Alessio Mora} received his Ph.D. in Computer Science and Engineering, at the University of Bologna, Bologna, Italy. He is currently a post-doctoral researcher at the University of Bologna. His research interests include Decentralized Learning, with a particular focus on Federated Learning, Deep Learning, and Edge Intelligence.
\end{IEEEbiography}

\vspace{-34pt}
\begin{IEEEbiography}[{\includegraphics[width=1in,height=1.25in,clip,keepaspectratio]{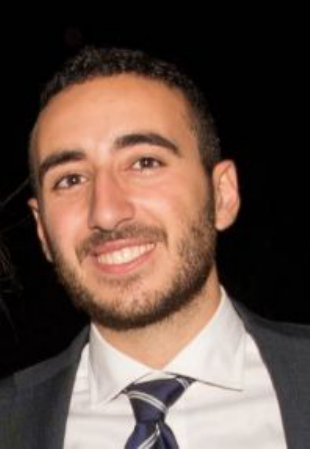}}]{Carlo Mazzocca}
received his Ph.D. in Computer Science and Engineering in 2024 from the University of Bologna, Bologna, Italy. Currently, he is an Assistant Professor at the University of Salerno, Salerno, Italy. His research primarily focuses on security and privacy aspects, with a particular emphasis on digital identity, security mechanisms built on distributed ledger technologies, authentication and authorization solutions for the cloud-to-thing continuum.
\end{IEEEbiography}

\vspace{-34pt}
\begin{IEEEbiography}[{\includegraphics[width=1in,height=1.25in,clip,keepaspectratio]{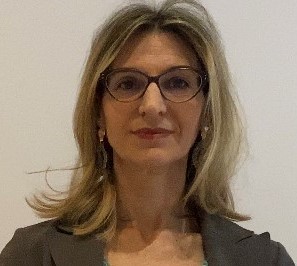}}]{Rebecca Montanari}
full professor at the University of Bologna since 2020 carries out her research in the area of information security and the design/development of middleware solutions for the provision of services in mobile and IoT systems. Her research is currently focused on blockchain technologies to support various supply chains, including agrifood, manufacturing and fashion, and on security systems for Industry 4.0.
\end{IEEEbiography}

\vspace{-34pt}
\begin{IEEEbiography}[{\includegraphics[width=1in,height=1.25in,clip,keepaspectratio]{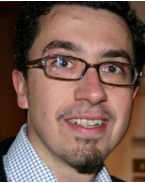}}]{Paolo Bellavista}
received the Ph.D. degree in computer science
engineering from the University of Bologna, Italy,
in 2001. He is currently a Full Professor with
the University of Bologna. His research interests include middleware for mobile computing,
QoS management in the cloud continuum, infrastructures for big data processing in industrial
environments, and performance optimization in
wide-scale and latency-sensitive deployment environments. He serves on the Editorial Boards of many international journals.
\end{IEEEbiography}

\newpage

\section*{Appendix - Experimental Setup}
\label{appendix:exp_setup}

\begin{figure}[t!]
\centering
\begin{subfigure}[t]{0.24\textwidth}
\centering
\includegraphics[width=.81\linewidth]{figures/mit-b0_test_vs_train_cifar100_500C_iid_kl.pdf}
\end{subfigure}
\hfill
 \begin{subfigure}[t]{0.24\textwidth}
 \centering
  \includegraphics[width=.81\linewidth]{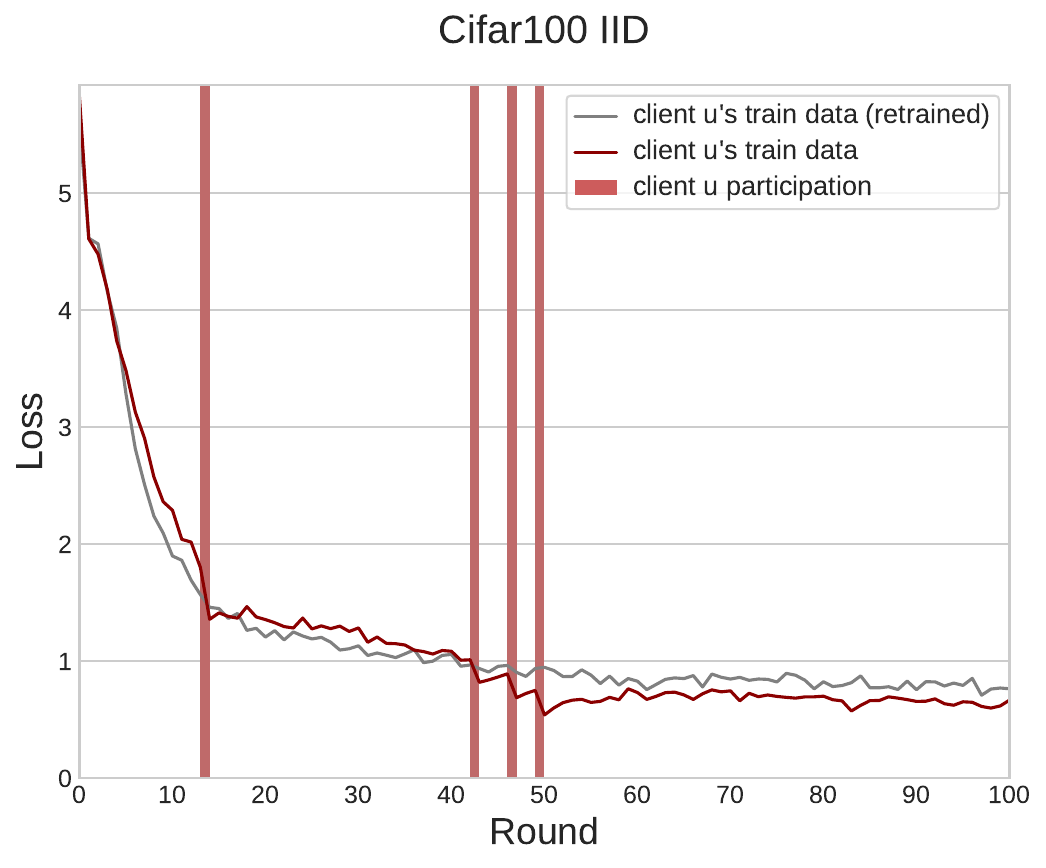}
 \end{subfigure}

  \begin{subfigure}[t]{0.24\textwidth}
 \centering
  \includegraphics[width=.81\linewidth]{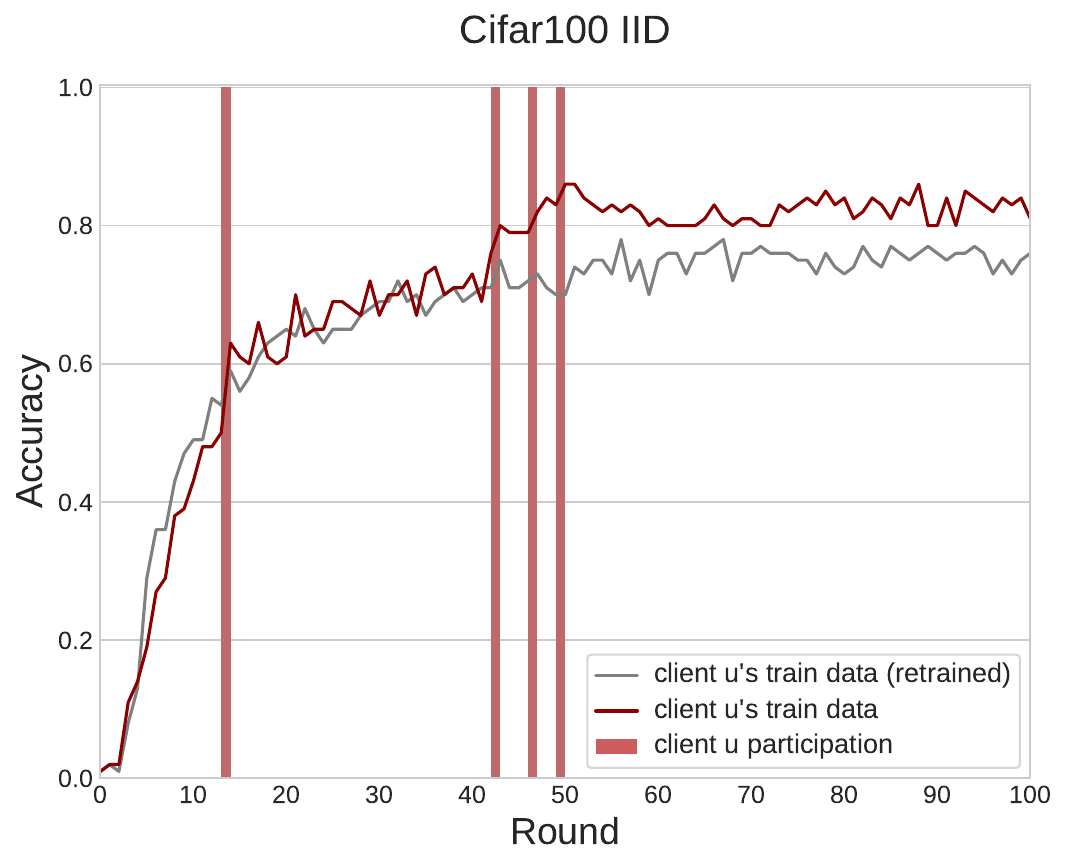}
 \end{subfigure}
 \hfill
\begin{subfigure}[t]{0.24\textwidth}
\centering
\includegraphics[width=.81\linewidth]{figures/mit-b0_test_vs_train_cifar100_500C_dir_0.1_kl.pdf}
\end{subfigure}

 \begin{subfigure}[t]{0.24\textwidth}
 \centering
  \includegraphics[width=.81\linewidth]{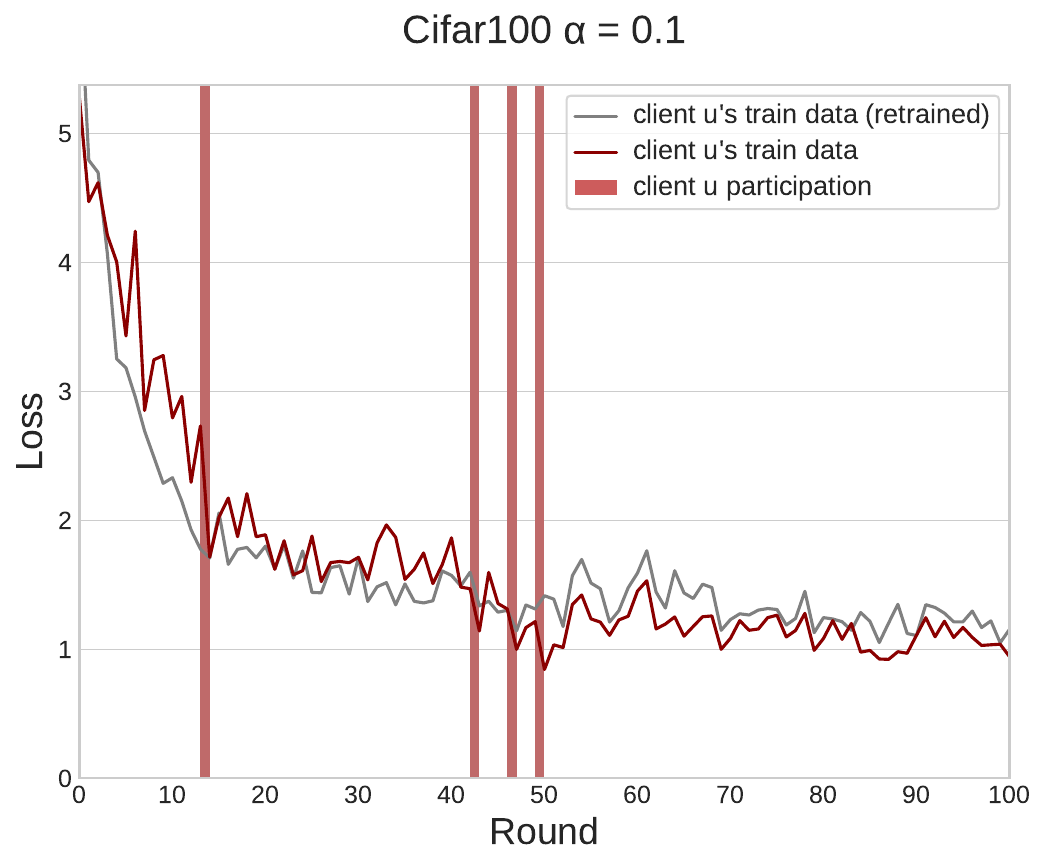}
 \end{subfigure}
 \hfill
  \begin{subfigure}[t]{0.24\textwidth}
 \centering
  \includegraphics[width=.81\linewidth]{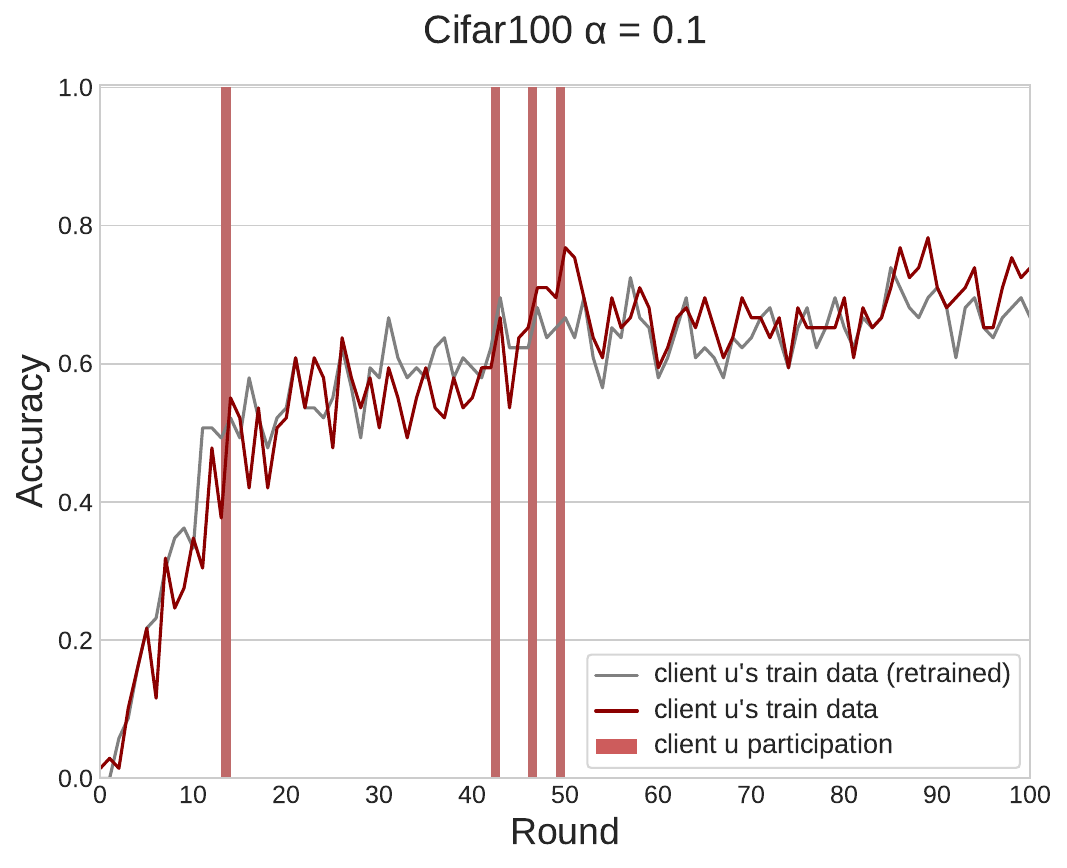}
 \end{subfigure}
\caption{KL divergence and accuracy for 100 rounds. 500 clients (4\% participation rate). Red line: client $u$ participates only during the first 50 rounds. Gray line: client $u$ never participates.}
\label{fig:natural_forgetting_complete}
\end{figure}

This section describes the experimental setting for the results reported in Table \ref{table:results_train_test}, Figure \ref{fig:test_vs_train_and_client_participation}, Figure \ref{fig:natural_forgetting}, and Figure \ref{fig:natural_forgetting_complete}. We implemented the simulations with Flower \cite{beutel2020flower} and TensorFlow. Our code is publicly available at \url{www.github.com/alessiomora/unlearning_fl}.

\subsection{Model Architecture and Hyperparameters}
\label{appendix:model}
We used a visual transformer, i.e., MiT-B0 \cite{xie2021segformer}, with approximately 3.6M parameters, initialized from a pre-trained model checkpoint trained on ImageNet-1k (69.27\% accuracy on test data). We adapted the one-layer classification head to this task, initializing such a layer from scratch. We employed the AdamW optimizer with a client learning rate of 3e-4, with a round-wise exponential decay of 0.998, 5 local epochs, batch size of 32, and weight decay regularization of 1e-3.

\subsection{Datasets}
We performed experiments on three datasets: CIFAR-100 \cite{Krizhevsky09learningmultiple}, Caltech-2011 (birds) \cite{WelinderEtal2010}, and FGVC-Aircraft (aircraft) \cite{maji13fine-grained}. We provided two settings for the data partitioning among clients: identically and independently distributed (IID) data, i.e., each client holds approximately the same number of per-class examples, and non-IID simulated via distribution-based label skew following the method from \cite{hsu2019measuring}. For the latter setting, we tuned the label skew using a concentration parameter of $  \alpha=0.1$ to rule the Dirichlet distribution.
The same data example is not repeated among multiple clients.

\smallskip
\noindent\textbf{CIFAR-100.} CIFAR-100 consists of 60,000 examples of 32x32 color images -- 50,000 for training and 10,000 for testing -- belonging to 100 classes. To match the transformer models' input size, we resized the images to a resolution of 224x224 pixels; we also preprocessed the training images with random crop and horizontal flip layers. For the experiments reported in Figure \ref{fig:test_vs_train_and_client_participation} and Table \ref{table:results_train_test}, we partitioned the training set to simulate 100 clients in the federation; we set 100 as the number of clients so that, in the IID setting, each client can have at least five per-class (unique) examples. At each round, 10 clients out of 100 were randomly selected to participate. For the experiments on \textit{natural forgetting}, reported in Figure \ref{fig:natural_forgetting} and \ref{fig:natural_forgetting_complete}, we partitioned the training set to simulate 500 clients in the federation, with 20 clients selected per round.

\smallskip
\noindent\textbf{Caltech-2011.} Caltech 2011 (birds) consists of 11,788 examples of color images -- 5,994 for training and 5,794 for testing -- belonging to 200 classes. To match the input size of the transformer models we resized the images to a resolution of 224x224 pixels; we also preprocessed the training images with random crop and horizontal flip layers, similarly to the work in \cite{hu2023federated}. We partitioned the training set to simulate 29 clients; we set 29 as the number of clients so that, in the IID setting, each client can have at least one per-class (unique) example. At each round, 5 clients out of 29 were selected.

\begin{table}[t!]

\begin{center}
{
\caption{Dataset-specific configurations.}
\label{table:config_simulations}
\begin{tabular}{l c c c c}
\toprule
Dataset & Classes & Train examples & Tot. Clients & Active Clients\\
\midrule
Cifar-100 & 100 & 50,000 & 100 - 500 & 10 - 20\\
Birds & 200 & 5,994 & 29 & 5\\
Aircraft & 100 & 6,667 & 65 & 7\\
\bottomrule
\end{tabular}
}
\end{center}
\end{table}

\smallskip
\noindent\textbf{FGVC-Aircraft.} The FGVC-Aircraft dataset contains 10,000 images of aircraft -- 6,667 for training and validation and 3,333 for testing. The aircraft labels are organized in a four-level hierarchy. We consider a classification at the variant level (e.g., Boeing 737-700). A variant collapses all the models that are visually indistinguishable into one class. The dataset contains 100 images for each of the 100 different aircraft model variants. We removed the copyright banner from the images by cutting off the bottom 20 pixels in height. To match the input size of the transformer models, we resized the images to a resolution of 224x224 pixels. We also preprocessed the training images with random crop and horizontal flip layers, similarly to \cite{hu2023federated}. We partitioned the training set to simulate 65 clients; we set 65 as the number of clients so that, in the IID setting, each client can have at least one per-class (unique) example. At each round, 7 clients out of 65 were randomly selected to participate. Table \ref{table:config_simulations} summarizes the dataset-specific configurations.

\subsection{Metrics}
We tracked accuracy and KL divergence. Accuracy is measured either on test data or on client $u$'s train data. The KL divergence is computed between the output distribution of the model on client $u$'s data and a uniformly distributed output probability. This metric should work as a proxy for the meaningfulness of model predictions (the greater the divergence the more meaningful the predictions).

\end{document}